\documentclass{article}
\usepackage{arxiv}
\usepackage{hyperref}
\hypersetup{
  pdfinfo={
    Title={Image Collage on Arbitrary Shape via Shape-Aware Slicing and Optimization},
  }
}

\usepackage[utf8]{inputenc} 
\usepackage[T1]{fontenc}    
\usepackage{hyperref}       
\usepackage{url}            
\usepackage{booktabs}       
\usepackage{amsfonts}       
\usepackage{nicefrac}       
\usepackage{microtype}      
\usepackage{lipsum}
\usepackage{graphicx}
\graphicspath{ {./images/} }
\usepackage{amsmath}
\usepackage[numbers]{natbib}
\usepackage{dirtytalk}

\usepackage{xcolor,soul,framed} 
\usepackage{enumitem}
\usepackage[ruled,linesnumbered]{algorithm2e}
\usepackage{wrapfig}

\usepackage{array}
\usepackage{float}
\usepackage{mdwmath}
\usepackage{amsmath}
\usepackage{algpseudocode}
\usepackage{mdwtab}
\usepackage{eqparbox}
\usepackage{url}
\usepackage{csvsimple}
\usepackage{amssymb}
\usepackage{multicol}
\usepackage{multirow, tabularx}
\usepackage{adjustbox}
\usepackage{makecell}
\usepackage{siunitx}
\usepackage{caption}
\usepackage{subcaption}

\DeclareMathOperator*{\argmin}{arg\,min}
\DeclareMathOperator*{\argmax}{arg\,max}

\title{Image Collage on Arbitrary Shape via Shape-Aware Slicing and Optimization}

\author{
 Dong-Yi Wu\\
 National Cheng-Kung University\\
  Taiwan\\
  \texttt{cutechubbit@gmail.com}
   \And
 Thi-Ngoc-Hanh Le \\
 National Cheng-Kung University\\
  Taiwan\\
  \texttt{ngochanh.le1987@gmail.com}
  \And
 Sheng-Yi Yao \\
 National Cheng-Kung University\\
  Taiwan\\
  \texttt{nd8081018@gs.ncku.edu.tw}
    \And
 Yun-Chen Lin \\
 National Cheng-Kung University\\
  Taiwan\\
  \texttt{f74042060@gmail.com}
  \And
  Tong-Yee Lee*\\  
  National Cheng-Kung University\\
  Taiwan\\
  \texttt{tonylee@mail.ncku.edu.tw}
}

\begin{document}
\maketitle

\begin{abstract}
Image collage is a very useful tool for visualizing an image collection. Most of the existing methods and commercial applications for generating image collages are designed on simple shapes, such as rectangular and circular layouts. This greatly limits the use of image collages in some artistic and creative settings. Although there are some methods that can generate irregularly-shaped image collages, they often suffer from severe image overlapping and excessive blank space. This prevents such methods from being effective information communication tools. In this paper, we present a shape slicing algorithm and an optimization scheme that can create image collages of arbitrary shapes in an informative and visually pleasing manner given an input shape and an image collection. To overcome the challenge of irregular shapes, we propose a novel algorithm, called \emph{Shape-Aware Slicing}, which partitions the input shape into cells based on medial axis and binary slicing tree. \emph{Shape-Aware Slicing}, which is designed specifically for irregular shapes, takes human perception and shape structure into account to generate visually pleasing partitions. Then, the layout is optimized by analyzing input images with the goal of maximizing the total salient regions of the images. To evaluate our method, we conduct extensive experiments and compare our results against previous work. The evaluations show that our proposed algorithm can efficiently arrange image collections on irregular shapes and create visually superior results than prior work and existing commercial tools.

\keywords{Image collection visualization, image collage, irregular shape layout}
\end{abstract}

\section{Introduction}
\begin{figure}
  \centering
  \includegraphics[width=0.65\linewidth]{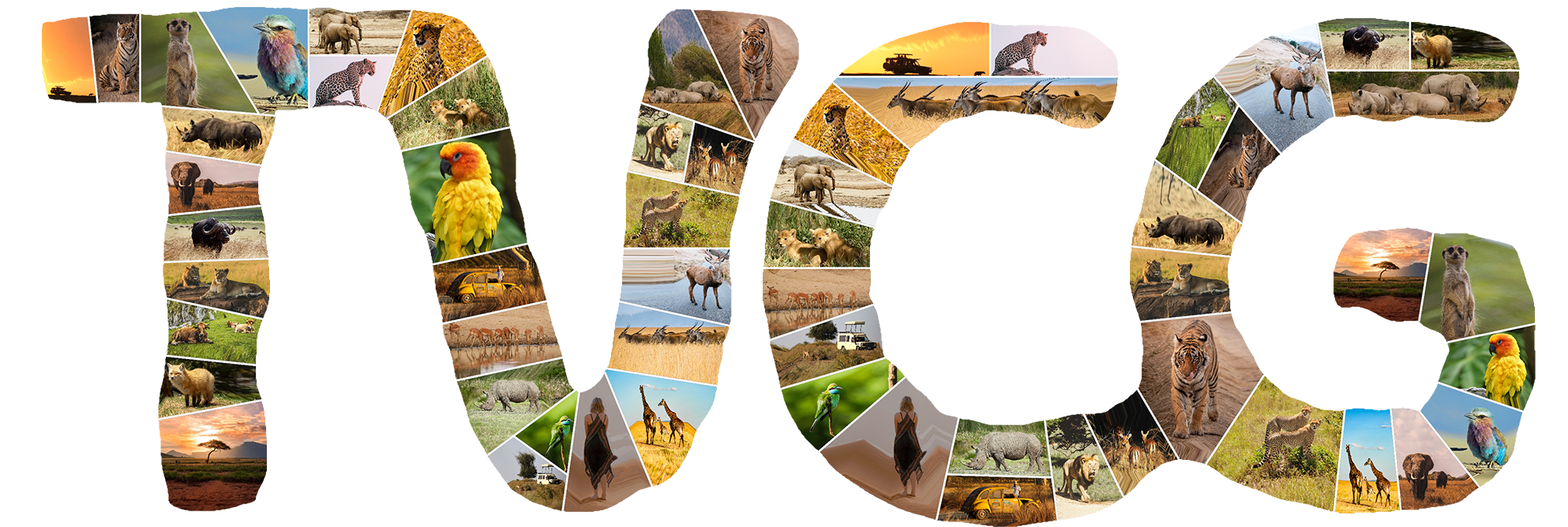}
  \caption{Images collages are generated by our proposed method. With the same image collection, we collage it to various irregular shapes.}
  \label{fig:teaser}
\end{figure}
\textcolor{black}{Image is mentioned as the way people use to visualize what they want to share via mobile devices. With the evolution of social media platforms (e.g., \textit{Twitter, Instagram, Facebook, Google Photos}, etc.), the need to share photos has become more attractive. An interesting way to visualize a photo collection is to collage them in an interesting or meaningful layout. The results may also be the way people use to represent the visual summary of their image collection with different purposes, for example, broadcast advertising (e.g., using the shape of a Kangaroo to visualize a collection of scenes in Australia), commemorating (e.g., using the shape of a heart to visualize a collection of wedding scenes). Such research domain is called in terms \textit{image collage}.}

\textcolor{black}{This exciting research topic has been studied early by various approaches. Researchers in \cite{wu2013picwall, liu2017trcollage, pan2019content, yu2022softcollage} focus on preserving the original aspect ratios of each image and missing the image content. Other approaches \cite{rother2006autocollage, tan2011imagehive, liu2017correlation} consider the content of images by trying to fit only the salient cutouts of each image into the canvas as fully as possible. In other words, their systems can generate a collage by overlapping images without occluding salient regions. However, most of these prior studies share the same difficulty in collaging image collection to an arbitrary shape. That is, they are all restricted to rectangle layouts. }

\begin{figure}
\centering
\begin{subfigure}[t]{0.22\textwidth}
  \centering
  \includegraphics[width=\textwidth]{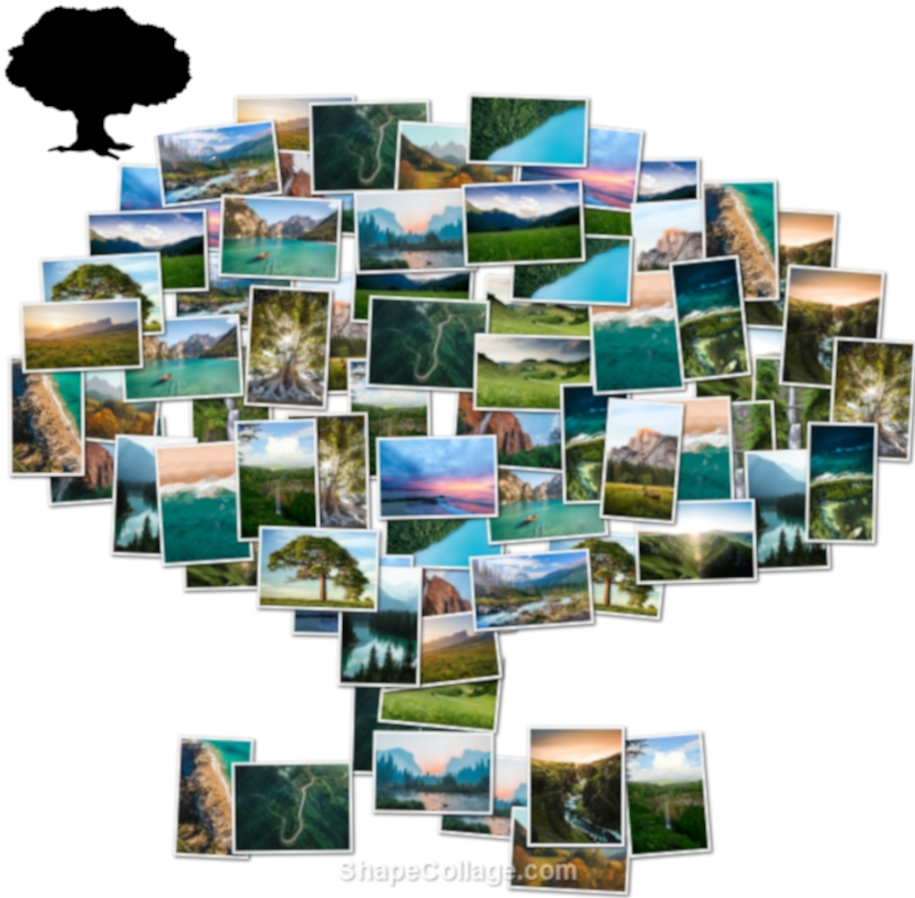}
  \caption{Shape Collage \cite{shapecollage}}
  \label{fig:shapecollage}
\end{subfigure}
\begin{subfigure}[t]{0.22\textwidth}
  \centering
  \includegraphics[width=\textwidth]{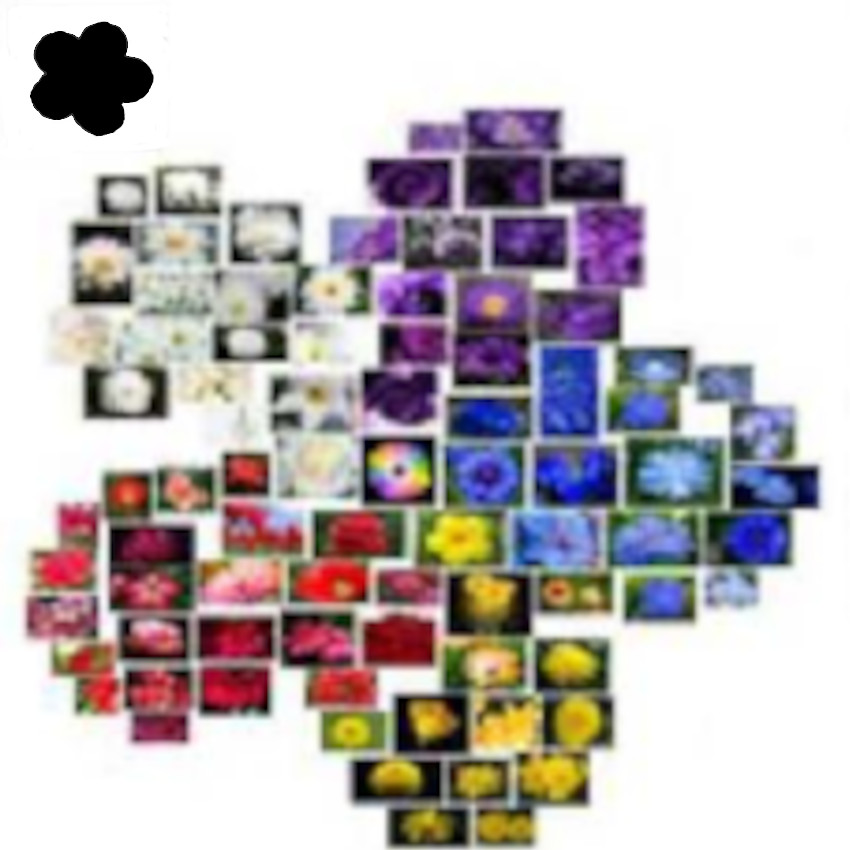}
  \caption{\citet{han2015tree}}
  \label{fig:treecollage}
\end{subfigure}%
\begin{subfigure}[t]{0.22\textwidth}
  \centering
  \includegraphics[width=\textwidth]{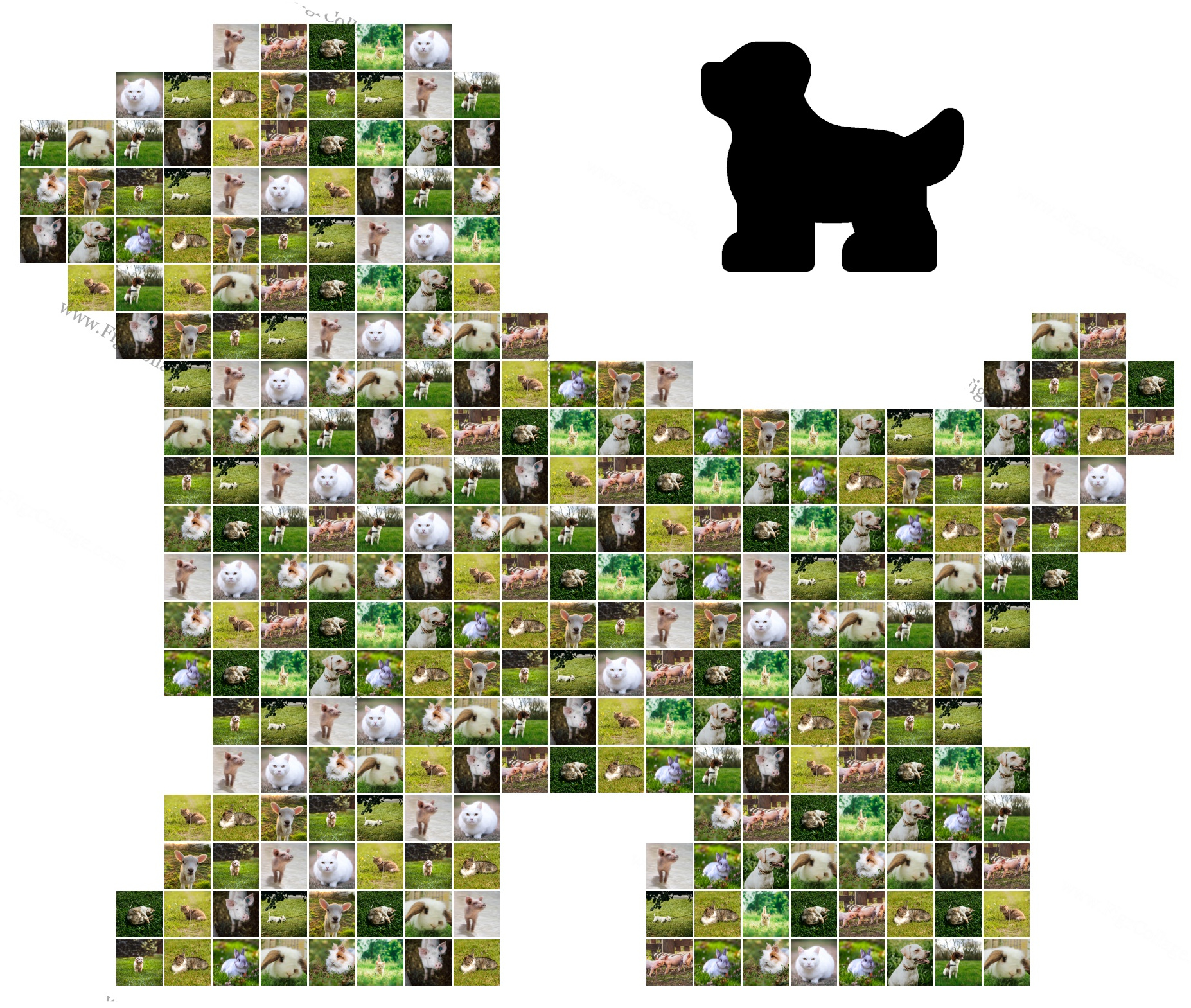}
  \caption{FigrCollage \cite{figrcollage}}
  \label{fig:figrcollage}
\end{subfigure}%
\begin{subfigure}[t]{0.22\textwidth}
  \centering
  \includegraphics[width=\textwidth]{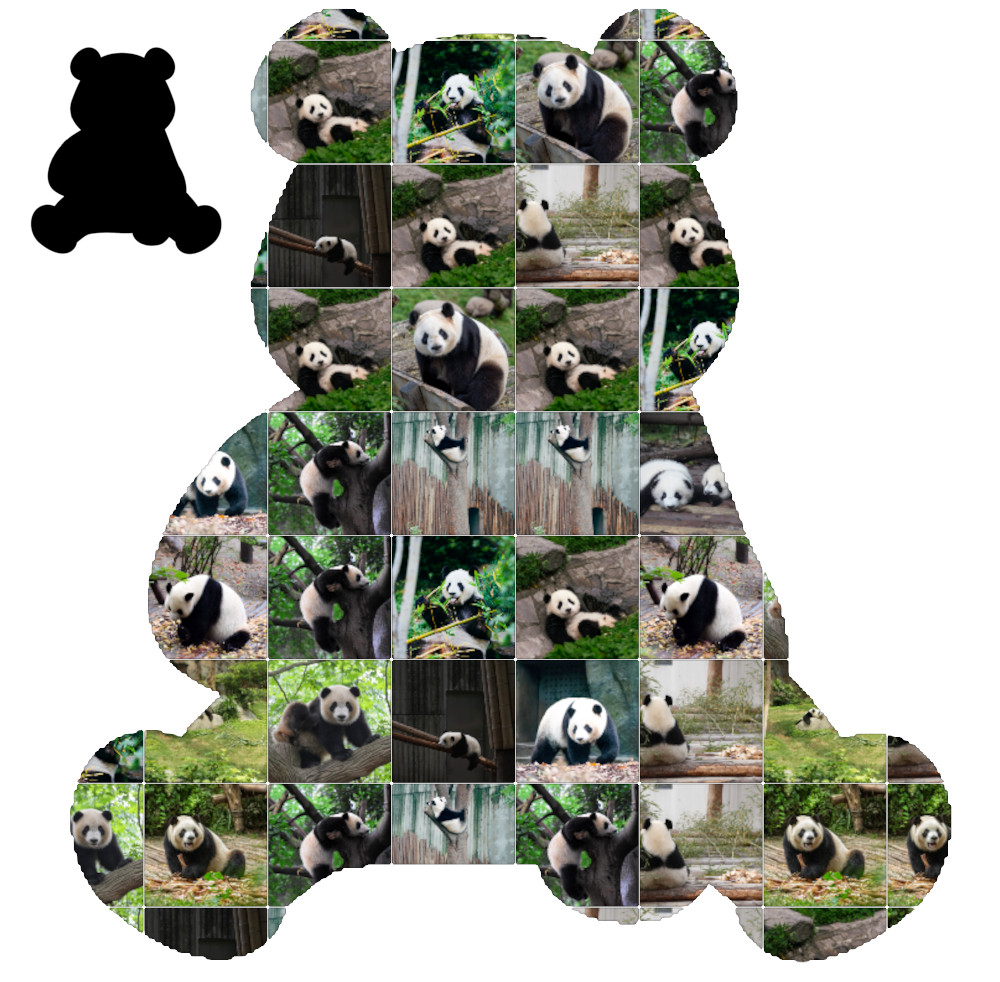}
  \caption{ShapeX \cite{shapex}}
  \label{fig:shapex}
\end{subfigure}

\caption{Example collages are generated by previous work and commercial applications. The black silhouette above the collage is the input shape.}
\label{fig:shapedimagecollage}
\end{figure}

Besides the above approaches, some commercial applications have been released for image collage in recent years, such as Shape Collage \cite{shapecollage}, FigrCollage \cite{figrcollage}, ShapeX \cite{shapex}, and Adobe \cite{Adobe}. With these applications, without any design experience necessary, people can craft their very own collage and allow their creativity to bring all their beautiful memories together. Nevertheless, they still suffer from some limitations. The images in resultant collages are heavily occluded \cite{shapecollage}. The cells in the generated layout are too small and uniform (e.g., rectangles or squares of the same size) \cite{figrcollage}. This issue makes the method face a fundamental trade-off between the image size and the accuracy of the layout shape. That is, images in the collection may have to be scaled down significantly to fully fit the layout. This phenomenon leads to that the collages are not visually pleasing. In ShapeX \cite{shapex}, a uniform grid is overlayed on the input shape without considering the shape structure. Hence, the collage generated by this application not only shares the same drawback with Shape Collage \cite{shapecollage} and ShapeX \cite{shapex} but also yields unpleasing regions at the boundary regions. \citet{han2015tree} attempt to collage on an irregularly shaped layout by first projecting images onto a 2D circular region and locally moving images within the target region. Hence, their method is designed to work for shapes that are not far away from a circle, e.g. a heart or an apple. This method is not designed for highly irregular shapes (e.g. a shape with a hole in the middle). These results are shown in Fig.\ref{fig:shapedimagecollage}.

This paper addresses the above problems and proposes an image collage on an arbitrary shape (abbreviated as \textbf{ICAS}) method, as shown in Fig.\ref{fig:teaser}. Our collaging technique considers both the input shape and the content information of the images in the given collection. This enables our method to be capable of generating visually pleasing collages. To achieve that, we propose an algorithm based on binary slicing 
trees, which shoulders the task of portioning the input shape into regions. To serve the visually pleasing collage, we define the subjects of images by an Image Content Analyzing process prior to collaging. To evaluate the effectiveness of our image collage approach, we test it with diverse input shapes and image collection. Appealing results are obtained from our evaluated experiments. We further compare our results to those of the previous works and existing commercial applications to demonstrate the advantage of our proposed framework.

\textcolor{black}{Our contributions are summarized as follows: 
\begin{itemize}
    \item We propose a novel ICAS algorithm.
    \item We develop a layout generation method, \textit{Shape-aware Slicing}, that is especially useful to deal with the convex-concave surface of irregular shapes. 
    \item The \textit{optimization procedure} we investigate in this current work enables such an image collage method to build a bridge between input shape, layout design, and visual content of image collection.
    \item Various experiments with shapes and image collections demonstrate that our method is more accessible and can produce more appealing results. This allows ordinary users to be easier to visualize their beautiful memories together.
\end{itemize}}

\section{Related Work}
\subsection{Image Collage}
We have already seen that the image collage methods can be categorized as rectangular and non-rectangular or content-aware and not content-aware. Another way to look at these works is how they arrange the images. Many works group images of similar content and place them in close proximity. \citet{liu2017correlation} use t-SNE to embed each image onto a 2D canvas based on the feature vectors.  \citet{tan2011imagehive} cluster images based on the correlation between images with the k-means algorithm and put them inside the same cell. \citet{pan2019content} consider the importance and aesthetics of the image when placing the images, where important images are placed closer to the collage center. \citet{song2021balance} emphasize the use of the overall compositional balance of the collage and arrange the image according to the balance-ware metrics. Some works focus on image summary capability, in which representative images are selected first from a large collection of images and then visualized. \citet{rother2006autocollage} select top-ranking images according to their representativeness, importance, and object location. \citet{pan2019content} greedily select images considering conciseness, diversity, and aesthetics. The latest work \cite{yu2022softcollage} proposes an innovative continuous tree representation to partition the canvas. This enables an end-to-end collage generation model to be trained with backpropagation. This formulation, however, can only be defined on rectangular canvases. Another line of work focuses on interactive visualization of collections of images. \citet{nguyen2008interactive} present a visualization scheme for more than 10,000 images. \citet{lekschas2020generic} propose a framework for visualizing and exploring small multiples including large image collections. 

In comparison with existing methods, our new method can be summarized as non-rectangular and content-aware. Salient objects will be preserved and placed according to shape structure. Important images will be placed at the most salient location.

\subsection{Shape Decomposition}

Planar shape decomposition methods can be broadly categorized into two classes. One tries to decompose shapes into convex polygons. The other attempts to mimic how humans partition a shape based on cognition research.

Earlier works \cite{latecki1999convexity, lien2006approximate} usually focus on decomposing shapes into convex parts. Conventional strict convex decomposition is a well-studied problem, but it is not directly applicable to most shape decomposition tasks. One of the shortcomings is that it will produce overly-segmented parts. \citet{latecki1999convexity} observe the phenomenon that non-convexity smaller than a certain scale is irrelevant to how humans perceive a shape. Thus, they develop the DCE algorithm to control the tolerance level of non-convexity. \citet{lien2006approximate} propose \emph{Approximate Convex Decomposition}, which decomposes shapes into approximately convex parts. We do not use \emph{Approximate Convex Decomposition} in this work, because, it will produce tiny partitions, which are not suitable for collage generation.

Later researches on shape decomposition attempt to develop computational models based on psychophysical findings. The most recognized rules derived from those findings are the \emph{minima rule} \cite{hoffman1984parts}, the \emph{short-cut rule} \cite{singh1999parsing}, along with the definition of \emph{part-cuts}\cite{singh2001part}. \citet{luo2014computational} propose an optimization model that realizes the aforementioned rules. \citet{papanelopoulos2019revisiting} make effective use of medial axis representation and capture most of the rules and saliency measures suggested by psychophysical studies, including the minima and short-cut rules, convexity, and symmetry. \citet{papanelopoulos2019revisiting}'s work, referred to as \emph{MAD}, is grounded in rigorous mathematical reasoning instead of relying heavily on heuristic rules like earlier methods. As a result, it is easier for us to adapt it for our own goal, in this case, generating image collages. Furthermore, it does not require complex optimization processes like the one in \citet{de2006segmentation}'s work and it achieves better performance in the public dataset than other works. \textcolor{black}{Our shape decomposition method utilizes this concept as the baseline to decompose the input shape into convex polygons. Thereafter, we investigate a novel slicing algorithm to generate the balanced and visually pleasing layout}.

\section{System Framework}
\begin{figure}
\centering
\includegraphics[width=0.8\textwidth]{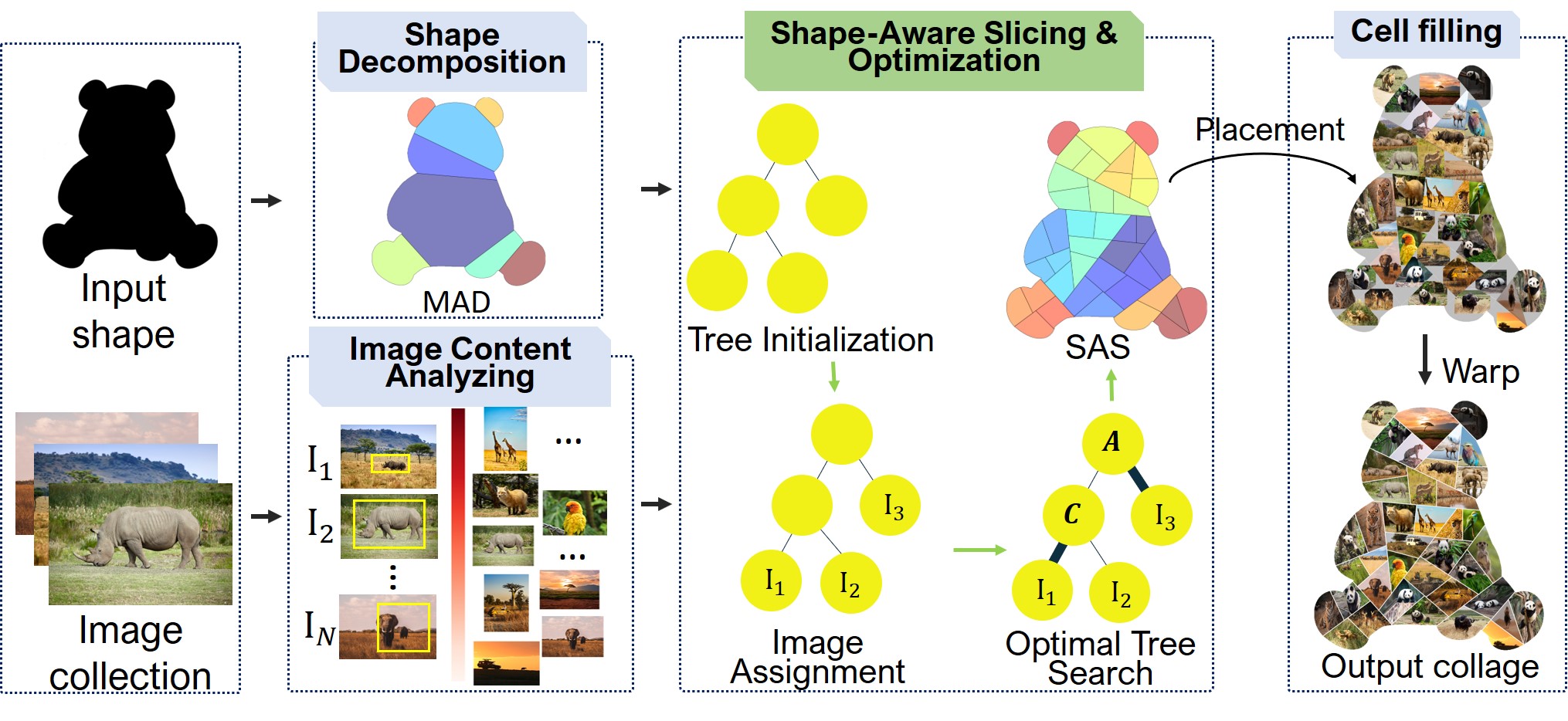}
\caption{System framework}
\label{fig:system_framework}
\end{figure}

The framework of our ICAS system is illustrated in Fig.\ref{fig:system_framework}, which consists of three main processes: \textit{Image Content Analyzing, Shape Decomposition}, and \textit{Shape-aware slicing and Optimization}. The proposed scheme takes as input an arbitrary shape and an image collection. Our goal is to generate an information-rich and beautifully-arranged shaped image collage.

\textbf{Image Content Analyzing} is proposed to define the important information of images before placing them in the layout. This process saves the resultant collage from poor aesthetics, e.g., the important objects are cropped out. This phenomenon is mentioned as a drawback in previous approaches \cite{shapecollage, han2015tree, shapex, figrcollage}. In our approach, the image collection is first passed through a salient object detection model. Accordingly, each image is associated with an importance score.

\textbf{Shape Decomposition} shoulders the task of portioning the highly irregular shape into regions, which are convex polygons. As we discussed in the prior session, the layout in an arbitrary shape is challenging, and it is also the key difference between our current method and previous works.

\textbf{Shape-aware Slicing and Optimization} is the main process in our workflow. The shape is further partitioned such that each region corresponds to an image in the given collection. We achieve this by first proposing Medial axis-based Binary Slicing Tree (MABST) and Shape-Aware Slicing(SAS) operations as a new way to partition an irregularly-shaped canvas. Then, we optimally select an optimal layout that can maximize the important region of a given image collection. Finally, our customized image warping technique is applied to create the final collage.

\section{Methodology}

\subsection{Image Content Analyzing}
To build a bridge between the image content and the layout design, we analyze the content of images in the given collection. Analyzing the content of images in the given collection enables our system to understand the semantics of individual images and the visual topic of the collection. To analyze the content of images in the collection, we adopt a supervised salient object detection model \cite{pang2020multi}. The subject for each image is simplified as a salient box $Sb = [bx_1, by_1, bx_2, by_2]$ containing all the salient pixels. Such a box is used to represent the important region of an image. We choose a bounding box representation instead of using the saliency map directly because the maximization of a rectangle’s area inside a convex polygon can be solved efficiently with linear programming. As we will show in the coming section, we need to calculate this value multiple times when we are searching for the optimal layout.

A plus of our collage system is that we allow users to designate the photos in the collection they are most interested in. We take such photos into account when placing the collection in the layout. For this reason, we encourage the users to perceptually choose the photos that are dominant in the collection in terms of visually pleasing or aesthetic factors. We record the images designated by the users and assign them a high importance score. In the cases that the users do not choose, we adopt \textit{NIMA} \cite{talebi2018nima} to measure their aesthetic scores. As a result, with a given collection, we have a set $\mathbf{I} = \{I_i\}, i = 0, \dots, N_I$, $N_I$ is the number of input images. Each $I_i$ is a tuple of $\beta_i$ and $R^m_i$ which respectively denotes the image's index and importance rank. Note that only the portion of images in the salient box will be assessed since the images are usually not fully visible in the final collage.

Three major benefits can be gained from this analysis. The bounding boxes help us to find the tailored cell that could be fitted to the area of the important region in an image. Second, this saves the subject in the images from cropping. Third, ranking the photos according to aesthetic scores and integrating them with the layout serves semantic and visually pleasing collage results.

\subsection{Shape Decomposition with medial axis}
 Shape decomposition algorithm decomposes arbitrary input shapes into manageable pieces, i.e. convex parts. The decomposition is accomplished by determining a set of part-cuts defined as line segments that divide the shapes into pieces. We adopt the state-of-the-art shape decomposition algorithm based on the medial axis (so-called MAD), which is introduced in \citet{papanelopoulos2019revisiting}. Before diving into shape decomposition algorithm, we briefly overview the medial axis used in their approach. 

Given a planer shape \(\mathbf{X}\subset \mathbb{R}^2\), the distance map \(D(\mathbf{X}): \mathbb{R}^2 \mapsto \mathbb{R}\) is a function mapping each point \(z \in \mathbb{R}^2\) to 
\begin{equation}
    D(\mathbf{X})(z) = \inf_{x \in \partial \mathbf{X}}\Vert z-x \Vert,
\end{equation}
where \(\Vert \;\; \Vert\) denotes the \(l^2\)-norm. For \(z \in \mathbb{R}^2\), let
\begin{equation}
    \pi(z) = \{z \in \partial \mathbf{X} : \Vert z-x \Vert = D(\mathbf{X})(z)\}
\end{equation} be the set of points on the boundary at a minimal distance to $z$. This is called the projection set of $z$ on the boundary. Each \(x \in \pi(z)\) is called a projection of $z$.

\begin{figure}
\centering
  \includegraphics[width=0.80\textwidth]{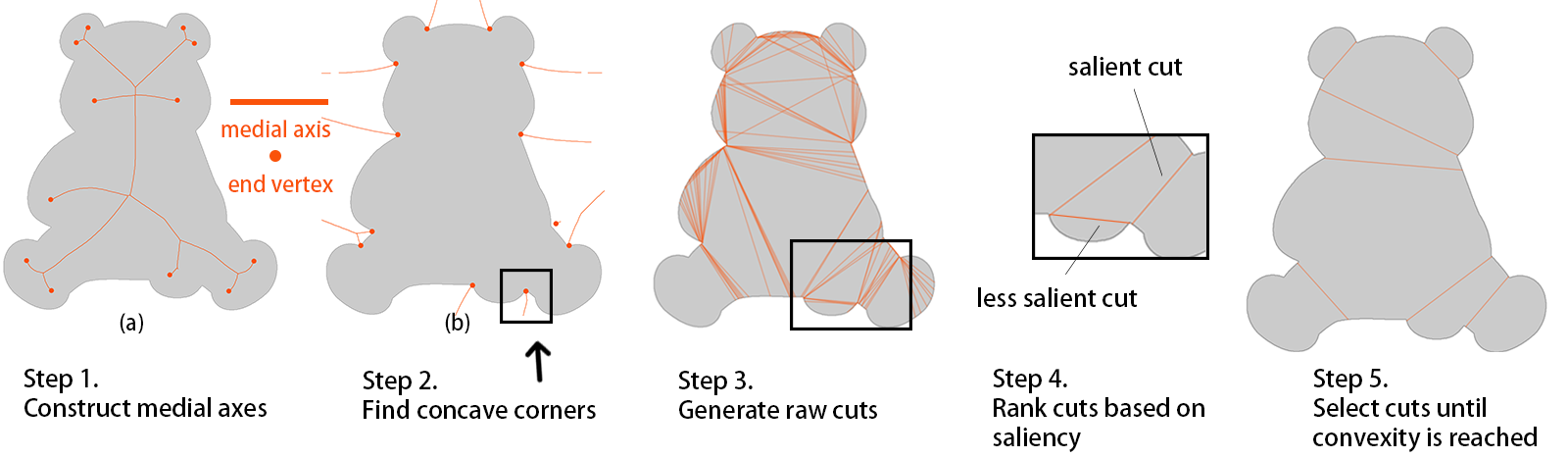}
\caption{Illustration of Step one through five of MAD. (a) and (b) are the interior medial axis and exterior medial axis.}
\label{fig:mad}
\end{figure}

The medial axis of shape $\mathbf{X}$ is a set of points of \(\mathbf{X}\) with more than one projection points, which is formulated as:
\begin{equation}
    M(\mathbf{X}) = \{z \in \mathbf{X}: \lvert \pi(z) \rvert > 1\}.
\end{equation}
This set can be interpreted as a finite linear graph in \(\mathbb{R}^2\) with the points that have exactly two projections as edges and others as vertices \cite{choi1997mathematical}. Fig.\ref{fig:mad} visualizes these mathematical definitions and the MAD algorithm step by step. A vertex is called as an \textit{end vertex} if it has degree one in the graph. Similarly, the exterior medial axis of \(\mathbf{X}\) can be defined as the medial axis of its complement \(\mathbb{R}^2 \setminus \mathbf{X}\). 

The medial axis carries information that is critical for decomposing \textcolor{black}{irregular} shapes. According to the minima rule \cite{hoffman1984parts}, part-cut endpoints should be the points of negative minima of curvature of the shape boundary, namely the concavity of the shape. It can be observed that the end vertices of interior (respectively to exterior) medial axis correspond to convex (respectively to concave) corners. More specifically, end vertices and their projections alone can determine the position, spatial extent, orientation and strength of the convexity (or concavity).

\begin{figure}
\centering
  \includegraphics[width=0.67\textwidth]{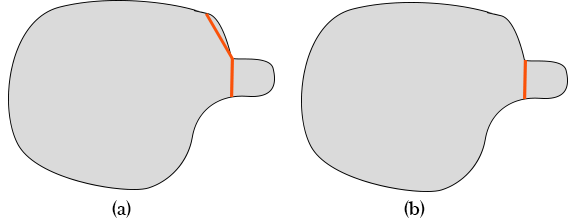}
\caption{(a) Before discarding cuts, (b) Discarding cuts that have higher protrusion strength}
\label{fig:ps}
\end{figure}

Once the concave corners are located (Step 2 in Fig.\ref{fig:mad}), part-cuts candidates can be formulated as line segments whose endpoints are the projections points of the interior medial axis and the starting point is the projection point in the concave corner (Step 3 in Fig.\ref{fig:mad}). We call this resultant part-cuts as \textit{raw cuts}. The raw cuts that humans are more sensitive to are prioritized (Step 4). Multiple measures are proposed to quantify the human sensitivity, i.e. protrusion strength, flatness, expansion strength and extension strength. Among them, the protrusion strength of a cut is the most critical metrics and is used in many other papers \cite{hoffman1997salience, zeng20082d}. It can affect the final appearance of our collage and is defined as the ratio of its length to the length of its corresponding arc along the boundary. In particular, the protrusion strength controls the level of details for our decomposition. The cuts that have protrusion strength greater than the threshold $\tau_p$ are discarded, as shown in Fig. \ref{fig:ps}. In all the examples in our paper, $\tau_p = 0.75$. In the final step (Step 5), the candidate cuts are selected greedily until convexity is achieved at every concave corner or all candidate cuts are selected. The final decomposition result is shown in Fig.\ref{fig:mad}, Step 5.

The goal of our method in our current application is to generate a balanced and visually pleasing layout with a defined number of cells. Thanks to MAD, we can control the significant convex-concave contours on shapes. However, to collage an image collection with diverse content and numerous images, MAD by itself is not sufficient to deal with these challenges. Thus, using MAD as the preprocessing step to initially decompose the input shape, we then seek a novel method to slice the decomposed parts to a satisfying layout. In the coming section, we present our approach to tackling this challenge.

\subsection{Shape-Aware Slicing}
The resultant parts obtained by MAD are convex polygons. We call each of them in the term \textit{\textbf{patch}}. We tackle the aforementioned challenge by proposing a new shape-aware slicing method that operates on each patch. Let $N_p$ be the number of patches, and $N_I$ be the number of images in the given collection. It is assumed that $N_I > N_p$. The method aims at portioning $N_p$ into cells ($N_c$) such that $N_c = N_I$. Our early experiments show that $N_I >> N_p$ in most cases. Yet, if the contrast cases occur, merging adjacent patches by itself is sufficient to yield a plausible layout.

Our slicing method is inspired by a strategy of floorplan design \cite{1586075}. This classical method is introduced for canvas partition based on slicing structure and a full binary tree. Such a slicing structure aims to recursively divide a rectangular canvas into smaller rectangles by horizontal splits and vertical splits. This strategy is then widely used to generate layouts in many image collage systems \cite{wu2013picwall, liu2017trcollage, pan2019content, yu2022softcollage}.

\begin{figure}
\centering
  \includegraphics[width=0.88\textwidth]{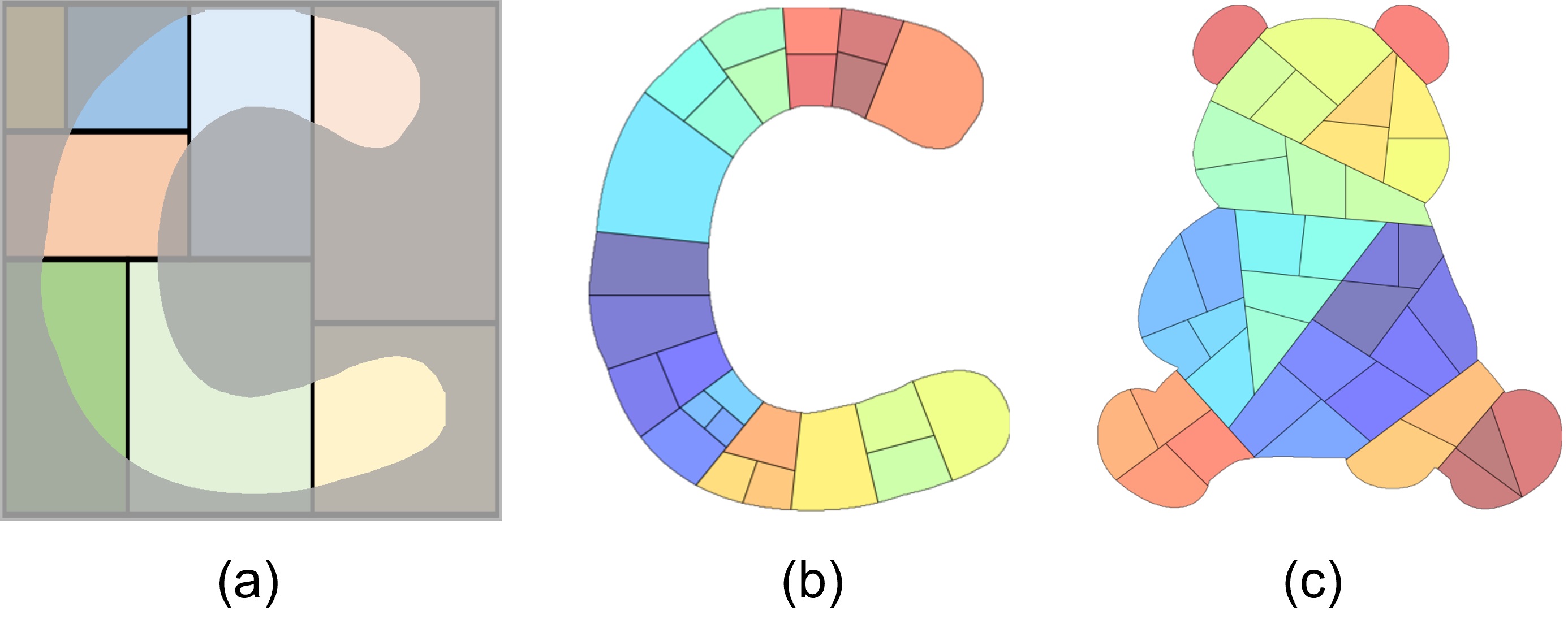}
\caption{Comparison on generated layouts. (a) layout by linear slicing, (b) and (c) are by our SAS algorithm.}
\label{fig:compare_layout}
\end{figure}

The challenge here is that we design the current system to handle various irregular shapes and orientations. Simply applying the slicing structure algorithm \cite{1586075} is insufficient, as the example in Fig.\ref{fig:compare_layout}-(a). Some cells are small, and some cells are not part of the shape. Hence, our SAS is designed differently from those in prior techniques \cite{atkins2008blocked, de2006segmentation,singh1999parsing}. It is observed that although the canvases are probably in various shapes and orientations, there is an intuitive \textit{horizontal} and \textit{vertical} direction. Such directions are relatively related to the medial axis concept, which we discussed in the earlier section. Therefore, we integrate the medial axis of the given shape to construct the binary tree, called Medial Axis-based Binary Slicing Tree (\textbf{MABST}).

\textcolor{black}{We utilize the medial axis of $\mathbf{X}$ to define the pseudo directions} that mimic the horizontal and vertical directions in a rectangular canvas\textcolor{black}{. They are respectively termed \textit{Axial} and \textit{Crosswise}}. For each point $z \in \mathbf{X}$, we define the closest point of $z$ in the medial axis set $M(\mathbf{X})$ as:
\begin{equation}
    \Phi(z,M(\mathbf{X})) = \argmin_{m \in M(\mathbf{X})}\Vert z-m \Vert
\end{equation}
\textcolor{black}{Accordingly, \(Axial\) and \(Crosswise\) of $z$ are defined as:}
\begin{itemize}
  \item \(Axial(z)\): The tangent vector of the medial axis at $\Phi\big(z\big)$. This is analogous to the horizontal cuts in the rectangular case. In practice, any one of the tangent vectors will suffice.
  \item \(Crosswise(z)\): A vector orthogonal to the Axial direction. This is analogous to the vertical cuts.
\end{itemize}

We visualize the pseudo directions in Fig.\ref{fig:axial_cr}-(a).

 Once the pseudo directions are defined, we initialize a MABST for each patch. We determine the number of images $\mathbf{S}$ to be assigned to a certain patch according to the patch's area. Given a shape $\mathbf{X}$ with a set of patches $\mathbf{P} = \{p_1, \dots, p_{N_p}\}$, we define $\mathbf{S}_i$ of patch $C_i$ as:
\begin{equation}
    \mathbf{S}_i = \big[N_I \cdot \dfrac{Area(p_i)}{Area(\mathbf{X})}\big], 
\end{equation}where \([ \;\;]\) is the notation of the nearest integer function.

For a MABST, each leaf represents a cell; and thus, the leaf count is the number of cells that matches $S_i$. Formally, a MABST is a recursive data structure. Each tree node $\mathbf{T}$, encompasses information of (1) cutting direction $\mathbf{D_\mathbf{T}}$ (Axial cut $\mathbf{A}$ and Crosswise cut $\mathbf{C}$), (2) the corresponding polygon $\mathbf{G_\mathbf{T}}$, (3) left child $\mathfrak{L}_{\mathbf{T}}$, and (4) right child $\mathfrak{R}_{\mathbf{T}}$. 

\begin{figure}
\centering
  \includegraphics[width=0.95\textwidth]{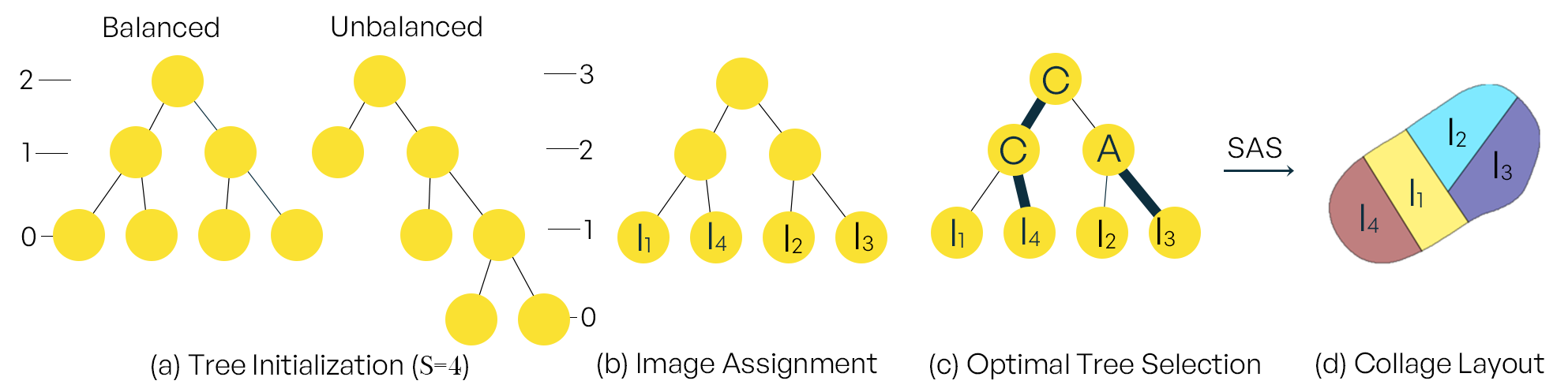}
\caption{The workflow of our layout generation process.}
\label{fig:layout}
\end{figure}

\begin{figure}
\centering
\begin{subfigure}{0.35\textwidth}
  \centering
  \includegraphics[width=\textwidth]{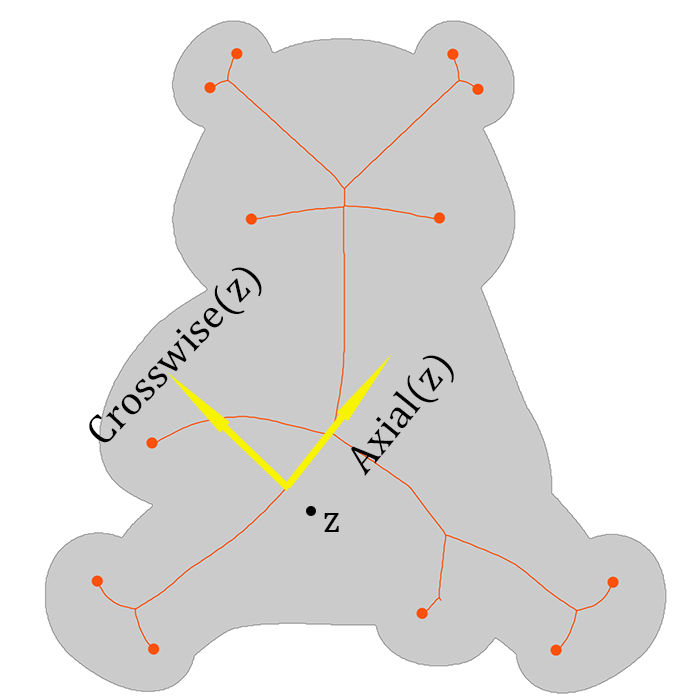}
  \caption{}
  \label{fig:ca}
\end{subfigure}
\begin{subfigure}{0.35\textwidth}
  \centering
  \includegraphics[width=\textwidth]{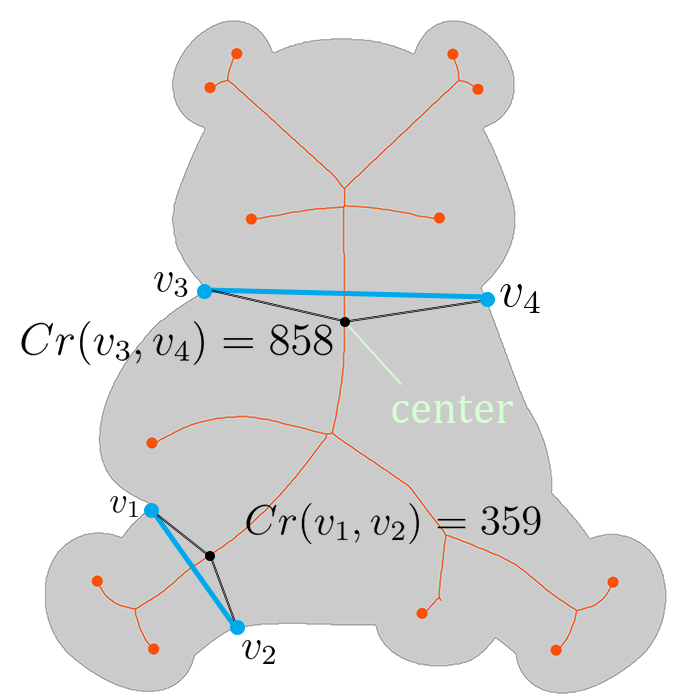}
  \caption{}
  \label{fig:cr}
\end{subfigure}%
\caption{(a) visualization of pseudo directions (yellow arrows), and (b) visualization of the process of finding the \emph{center}. Black points are on medial axis. Projections of them are $v_1$, $v_2$, $v_3$, and $v_4$, respectively. It can be observed that projections of the \emph{center}, $v_3$ and $v_4$ have the maximum Chord residual. }
\label{fig:axial_cr}
\end{figure}

Besides the number of leaf nodes, we also consider the balance of the tree when initializing the \textbf{MABST}. Obviously, a balanced tree yields even-sized cells and vice versa. Uneven-sized cells could be used to place less important images, such as landscape images. In practice, a splitting command propagates from the root node to a leaf node and splits a leaf node into two new leaf nodes. We repeat this operation $S_i - 1$ times starting from a single node. We select a branch for the splitting command to propagate based on the probability of \textit{Balanced} ($\gamma^b$) and \textit{Unbalanced} ($\gamma^u$). $\gamma^b$ is to select the branches with the least height, i.e., the number of edges on the longest path from the tree’s root node to a leaf. Meanwhile, $\gamma^u$ is to select the branches with the biggest height. However, we do not always want the MABST to become a degenerate linear path. Hence, some randomness is added with greater probability (i.e., $70\%$ in our experiments) to choose the branch with the biggest height. Two examples of MABSTs in this stage are shown in Fig. \ref{fig:layout}(a). So far, the MABST is not fully configured, i.e. the cut direction and the image association is not yet decided. The process of assigning images to leaf nodes will be discussed in Section 4.4. In Section 4.5, we will discuss how to decide the cutting direction for each node.

The core of our layout generation is the Shape-aware slicing algorithm \textbf{SAS}. SAS maps a MABST to a 2D collage layout. Let's first assume that we have a fully-configured MABST. SAS recursively iterates through every node $T$ and divides the polygon $\mathbf{G}_T$ according to the cutting direction ($\mathbf{A}$ or $\mathbf{C}$) with the help of a function \emph{Dividing Polygon} \textbf{DPG}. DPG divides a polygon by cutting it in half with a line passing through the polygon's centroid with the slope determined by the \emph{Axial} or \emph{Crosswise} direction. Since the polygon is convex, we can be sure that the centroid is inside the polygon, and there are precisely two resultant polygons. After the SAS operation, we simply collect all the polygons from leaf nodes as our final layout. The pseudo-code of the SAS and DPG algorithm are presented in Algorithm \ref{alg:shapeawareslicing} and Algorithm \ref{alg:dividepolygon}, respectively. Note that in Line 7 and 8 of the SAS algorithm, there is an additional parameter that we need to decide, namely the order of two child nodes. We can assign the polygon $p_1$ to the left child and the polygon $p_2$ to the right child and vice versa. This results in two different collage layouts. This decision will also be discussed in Section 4.5.

\SetKwComment{Comment}{/* }{ */}
\SetKwInOut{kwInput}{Input} 
\SetKwInOut{kwOutput}{Output}
\SetKwInOut{kwInit}{Initialization}
\SetKwRepeat{kwRepeat}{Repeat}{}
\SetKwRepeat{kwUntil}{Until}{}
\begin{algorithm}[h!]
\caption{\textbf{SAS} function}
\label{alg:shapeawareslicing}
\SetKwFunction{FMain}{SAS}
\SetKwFunction{FSub}{DPG}
\SetKwProg{Fn}{Function}{:}{}
    \Fn{\FMain{$T$, $M(\mathbf{X})$}}{
    \kwInput{Tree node: $T$, Medial axis: $M(\mathbf{X})$}
    
    \uIf {$T$ is not a leaf}{
        
        \uIf {$D_T$ is $A$}{
            \Comment{A split}
            $p_1, p_2 \gets $\FSub{$\mathbf{G}_T$, $M(\mathbf{X})$, $A$}\; 
        }
        \uElse { \Comment{C split}
            $p_1, p_2 \gets $\FSub{$\mathbf{G}_T$, $M(\mathbf{X})$, $C$}\; 
        }
        $\mathfrak{L}_T$.$\mathbf{G} \gets p_1$ \;
        $\mathfrak{R}_T$.$\mathbf{G} \gets p_2$\;
        
        \FMain{$\mathfrak{L}_T$, $M(\mathbf{X})$}\;
        \FMain{$\mathfrak{R}_T$, $M(\mathbf{X})$}\;
    }
}
\end{algorithm}

\begin{algorithm}[H]
\caption{\textbf{DPG} function}
\label{alg:dividepolygon}
\SetKwFunction{FMain}{DPG}

\SetKwProg{Fn}{Function}{:}{}
    \Fn{\FMain{$G$, $M(\mathbf{X})$, $D$}}{
    \kwInput{Polygon: $G$, Medial axis: $M(\mathbf{X})$, Cutting direction: $D$}
    \kwOutput{Two polygons $p_1$, $p_2$ result from the division}
    \uIf {$D$ is \emph{A}}{
        $ct \gets G.centroid $\;
        $slope \gets Axial(ct)$\;
        $dividing\_line \gets $ a line pass through $ct$ with $slope$\;
        $p_1, p_2 \gets  G$ divided by $dividing\_line$\;
        \Return $p_1, p_2$
    }
    \Else{
        \Comment{$D$ is \emph{C}}
        $ct \gets  G.centroid $\;
        $slope \gets Crosswise(ct)$\;
        $dividing\_line \gets $ a line pass through $ct$ with $slope$\;
        $p_1, p_2 \gets  G$ divided by $dividing\_line$\;
        \Return $p_1, p_2$
    }
}
\end{algorithm}

We show two layouts generated by our SAS algorithm in two sample shapes (e.g., character \say{C} and Panda) in Fig.\ref{fig:compare_layout}. We can see that SAS performs much better than the classical slicing algorithm on the shape of character \say{C}. Especially, Panda is a challenging shape since it has a large convex-concave contour. Even so, SAS still yields a balanced and visually pleasing layout. In particular, the elements in the generated layout, so-called \textit{cells}, are divided relatively evenly, and the specific regions (e.g., the ears or the legs) are well sliced. More results and comparisons are exhibited in the later experimental result section.

So far, we have introduced the concept of MABST and the mapping from trees to layouts. $N_p$ MABSTs are initialized such that each has $S_i$ leaf nodes. Before we arrive at a final slicing tree, we need to take image property, i.e. aspect ratio, into consideration. We will discuss how we assign images to a MABST in Section 4.4.

\subsection{Image Assignment}

We consider two factors when we assign images to leaves of the MABSTs: (1) Leaf nodes that are higher up in the tree is larger. The images with higher importance score should be assigned to larger cells, which are more prominent. (2) The images with higher importance score should be placed closer to the center of the shape, which attracts humans' attention. For example, the ear and feet of the panda shape is less prominent. We implement this idea by ranking MABSTs in terms of their inverse distance to the shape's center.

Intuitively, leaf nodes that are higher up in the trees are larger. But to define which node is higher, we can not directly use the height definition of a tree node since every leaf node has height zero. Instead, we define a quantity called \emph{elevation} of a node, which is defined as the height of the whole tree minus the depth of that node. The numbers in Fig. \ref{fig:layout}(a) shows the elevation of the nodes in two trees.

The \emph{elevations} of the leaf nodes are compared across all MABSTs. We further rank leaf nodes with the same elevation by their corresponding patches' distance to the center. Determining the center of an arbitrary shape is not trivial. For example, the centroid of a shape is not necessarily inside the shape. Hence, we adopt Chord residual \cite{ogniewicz1992voronoi} to determine the \emph{center} of an arbitrary shape. Given a line segment within the shape connecting two points \(v_i\) and \(v_j\) on the shape boundary $\mathbf{B}$ (as shown in Fig.\ref{fig:axial_cr}-(b)), Chord residual of them is formulated as:
\begin{equation}
  CR(v_i, v_j) = dist^B(v_i, v_j) - Length(	\overline{v_i v_j}) ,  
\end{equation}
where \(dist^B\) denotes the distance along the boundary $\mathbf{B}$. Accordingly, given the medial axis of a shape, the \emph{center} of the shape is formulated as:
\begin{equation}
    center = \argmax_{m \in M(X)} \; CR(v_i, v_j) \;\mid v_i, v_j \in \pi(m),
\end{equation} where $\pi(m)$ is the projection set, which was discussed earlier in section 4.2. We note here that the Chord residuals decrease as we move away from the \emph{center} along the medial axis, as illustrated in Fig.\ref{fig:axial_cr}-(b).

The prominence of a certain patch $\mathbf{PP}$ (and correspondingly the prominence of a MABST) can be expressed in terms of the inverse of the distance to the \emph{center}. The distance term is the sum of two distances: (1) the centroid of the patch \(p_i\), denoted by $p^e_i$ to its projection on the medial axis \(\Phi(p^e_i)\), and (2) the distance along the medial axis from \emph{center} to \(\Phi(p^e_i)\). Formally written as:
 \begin{equation}
    \mathbf{PP}(p_i) = \frac{1}{Length(\overline{p^e_i \Phi(p^e_i)}) + dist^{M(X)}(center, \Phi(p^e_i)},
\end{equation}
where \(dist^{M(X)} \) denotes the distance along the medial axis of the shape \(\mathbf{X}\); $\Phi(.)$ is the function in Equation (4).

Given the image importance rank $R^m$, we greedily select a leaf node of the highest elevation from all MABSTs and assign the most important image to it. We break the tie with the patch prominence $\mathbf{PP}$. If elevation and patch prominence are equal, images are assigned sequentially from left to right. This will lead to images of similar importance being placed together, which can improve informativeness. The MABST with images assigned is illustrated in Fig. \ref{fig:layout}(b).

\subsection{Optimal Tree Search}
\begin{figure}
\centering
  \includegraphics[width=0.65\textwidth]{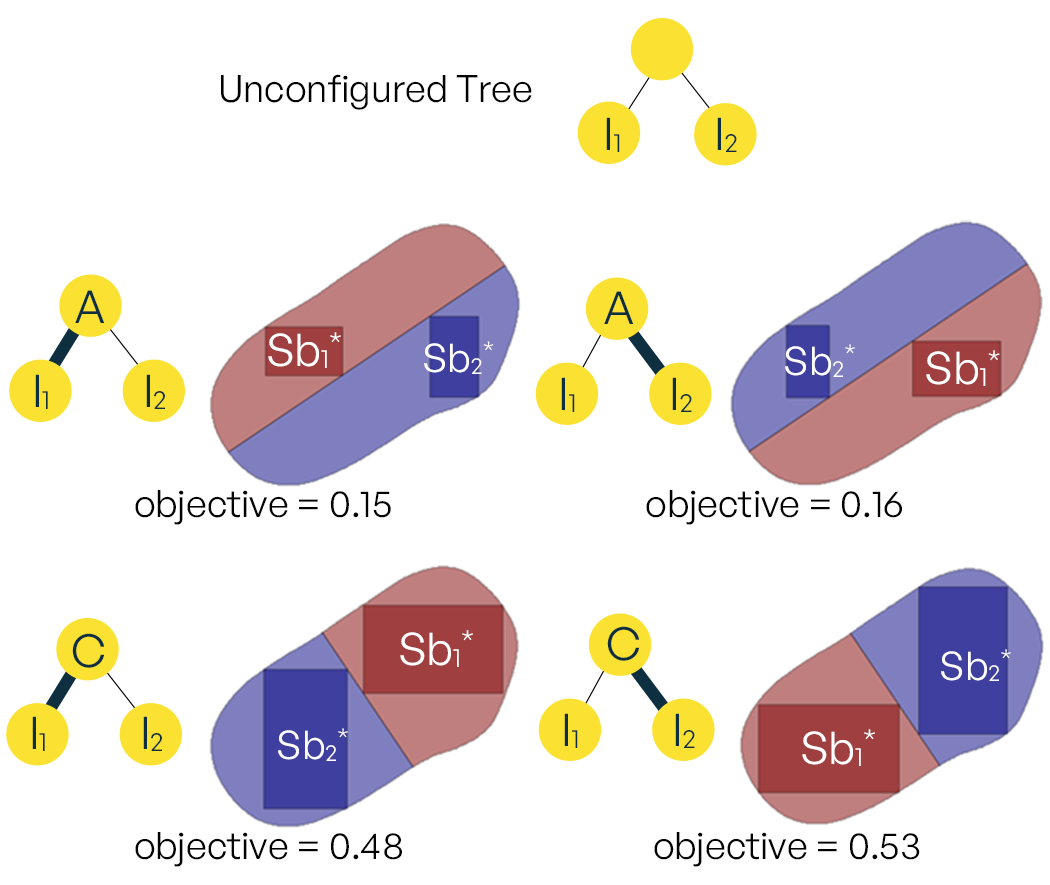}
\caption{The process of search for the optimal configuration for a tree. Four instances of the trees are shown alongside the corresponding layouts. $A$ and $C$ on the tree nodes represent the cutting direction and we use thicker edges to denote the larger polygons.}
\label{fig:optimization}
\end{figure}

We now find the optimal configuration for our slicing tree. The configuration $\mathbf{\mathcal{O}}$ for a tree $T$ refers to two things: (1) cutting direction $\mathbf{D_i}$ and (2) the order of two children $\mathbf{K_i}$ for $S_T-1$ inner nodes, where $S_T$ is the number of leaf nodes of $T$. We aim to find the layout structure that can maximize the total area of the maximum salient boxes $Sb^*$ of all images. The problem is illustrated in Fig. \ref{fig:optimization}. It can be seen that the bottom right layout in Fig.\ref{fig:optimization} has the largest objective value because two salient boxes are maximized. Formally, the optimization step determines an optimal configuration $\mathbf{\mathcal{O}}^{*}$
\begin{equation}
  \mathbf{\mathcal{O}}^{*} = \argmax_{\mathbf{D_i}, \mathbf{K_i}} \; E_{area},  
\end{equation}
where 

\begin{equation}
  E_{area} = \sum_{i=1}^{S_T} Area(Sb_i^*)
\end{equation}

Note that finding $Sb_i^*$ itself is an optimization problem. $Sb_i^*$ is defined as a rectangle of a maximum size that is fully inside a convex polygon and has the same aspect ratio as $Sb_i$, as shown in Fig. \ref{fig:optimization}. This problem can be efficiently solved using linear programming by representing convex polygons as the intersection of half-planes.

To find the optimal tree configuration $\mathbf{\mathcal{O}}^{*}$, we need to go through every possible configurations and find the best set of decision variables with the largest $E_{area}$. However, for a $S$-leaf-node tree, there are $4^{S-1}$ ways to configure the tree because each non-leaf node (inner node) has four possible configurations (Fig. \ref{fig:optimization}). In other words, the search space grows exponentially with the number of leaf nodes, which becomes intractable even for modest $S$.

We observe that nodes that are higher up in the tree correspond to rougher cuts in the final layout. This rougher cuts has less contribution to the final shapes of the leaf nodes, especially for every deep leaf nodes. For example, whether we select a Axial or a Crosswise cut for our first cut matters little when we intend to fit 50 cells inside this shape. Using this observation, we propose a simple strategy to reduce the search space by pre-configuring the inner nodes that have elevation higher than $\tau_e$, where $\tau_e$ is adjustable based on the trade-off of quality and speed. It is clear that the higher the $\tau_e$ the closer it is to the original brute-force search and vice versa.

\begin{figure}
\centering
  \includegraphics[width=0.8\textwidth]{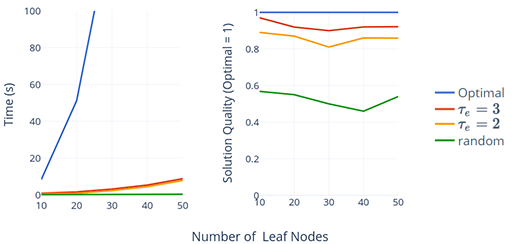}
\caption{The trade-off of execution time and solution quality with different searching strategies. Optimal indicates the use of brute force search.}
\label{fig:optimization_quality}
\end{figure}

To pre-configure the cutting direction for an inner node, we project the polygon associated with that node along the \textit{Axial} and \textit{Crosswise} axis and compare their dimension in these two directions. If the dimension in the \emph{Axial} axis is greater, a $\mathbf{C}$ cut is used. Otherwise, $\mathbf{A}$ cut is used. This is analogous to splitting a tall rectangle with a horizontal and dividing a wide rectangle with a vertical in the rectangular case. This prevents the resultant rectangles from having extreme aspect ratios, which may not be good for the quality of the cells. Fig.\ref{fig:optimization_quality} shows that using this strategy can greatly speed up the search time and achieve good objective values. From the experiments, setting $\tau_e = 3$ can consistently achieve more than 90\% of the optimal results for all leaf node counts, which is considerably better than fully random configuration (the green line in Fig.\ref{fig:optimization_quality}). All the results in latter part of this paper use this settings.

\textbf{Triangle Penalty.} The cells generated with SAS are usually quadrilaterals (except for cells on the boundary). But sometimes there will be triangles and these triangular cells tend to stand out from the rest of the shapes, which negatively impacts the uniformity of the cells. Consequently, we add a triangle penalty term $p_{triangle}$ to our objective function to discourage the optimization function from selecting triangular cells. We empirically set this penalty to 0.8 in our experiments to gain a balanced layout for arbitrary input shapes.
\begin{equation}
    p_{triangle}(polygon) = \begin{cases}
              0.8  & polygon \text{ is a triangle} \\
              1.0 & \text{otherwise.}
          \end{cases}
\end{equation}
We penalize its area term in the objective function by $p_{triangle}$:
\begin{equation}
  E_{area} = \sum_{i=1}^{S_T} (Area(Sb_i^*) \cdot p_{triangle}(\mathbf{G})),
\end{equation}
where $\mathbf{G}$ is the polygon associated with that leaf node.

For the time complexity, the brute-force search is $O(4^n)$. Using our strategy, we can reduce it to $O(n)$, which is verified by the linear trend in Fig. \ref{fig:optimization_quality}. For example, if we have a 8-leaf tree and we set $\tau_e = 1$, we only need to configure $8/2$ inner nodes that is immediately above the leaf nodes. Each inner node has 4 configurations. The total search space is $4^1\cdot8/2$ since these four nodes are independent. For $n$ leaf-node tree the search space is  $4^1\cdot n/2$. For $\tau_e = 2$, the number is $4^3\cdot n/4$. In general, the size of the search space is $4^{2^{\tau_e}-1}\cdot n/2^{\tau_e}$, which is linear in terms of $n$.

\subsection{Cell Filling}
\begin{figure}
\centering
\includegraphics[width=0.8\textwidth]{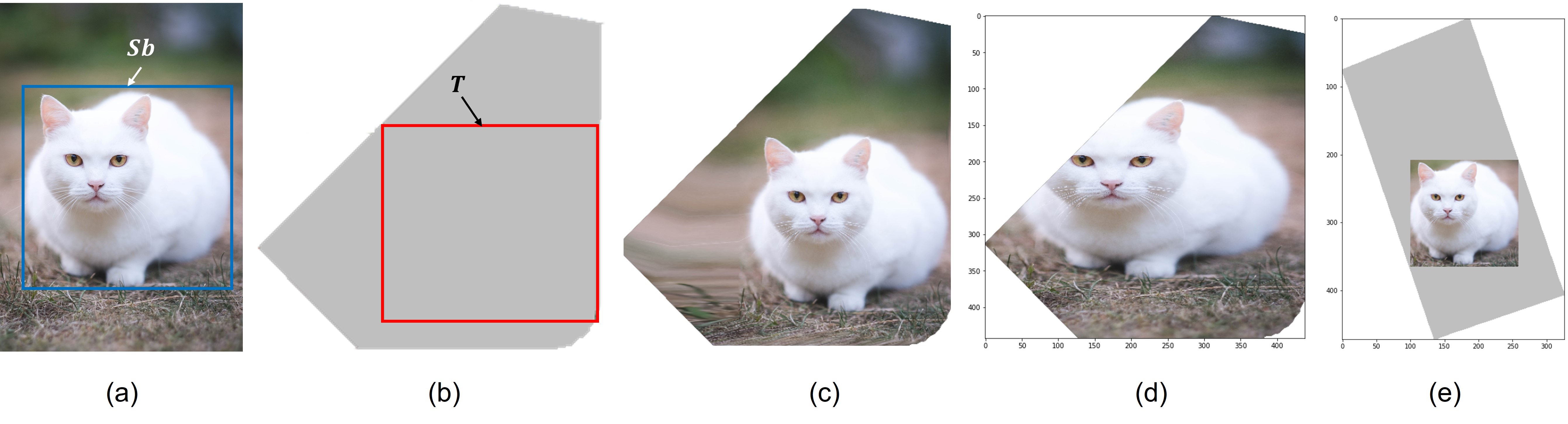}
\caption{\textcolor{black}{Results with optimization: (a) Image in the collection with the detected bounding box $Sb$ (blue rectangle); (b) estimated box $T$ in cell to make $Sb$ fit $T$; (c) Filling the cell by warping. (d) and (e) are failed results without optimization: object in image is cutout to fit the cell (d); fail to assign a tailored cell to the image (e).}}
\label{fig:optimize_process}
\end{figure}
\begin{figure}
\centering
\includegraphics[width=0.8\textwidth]{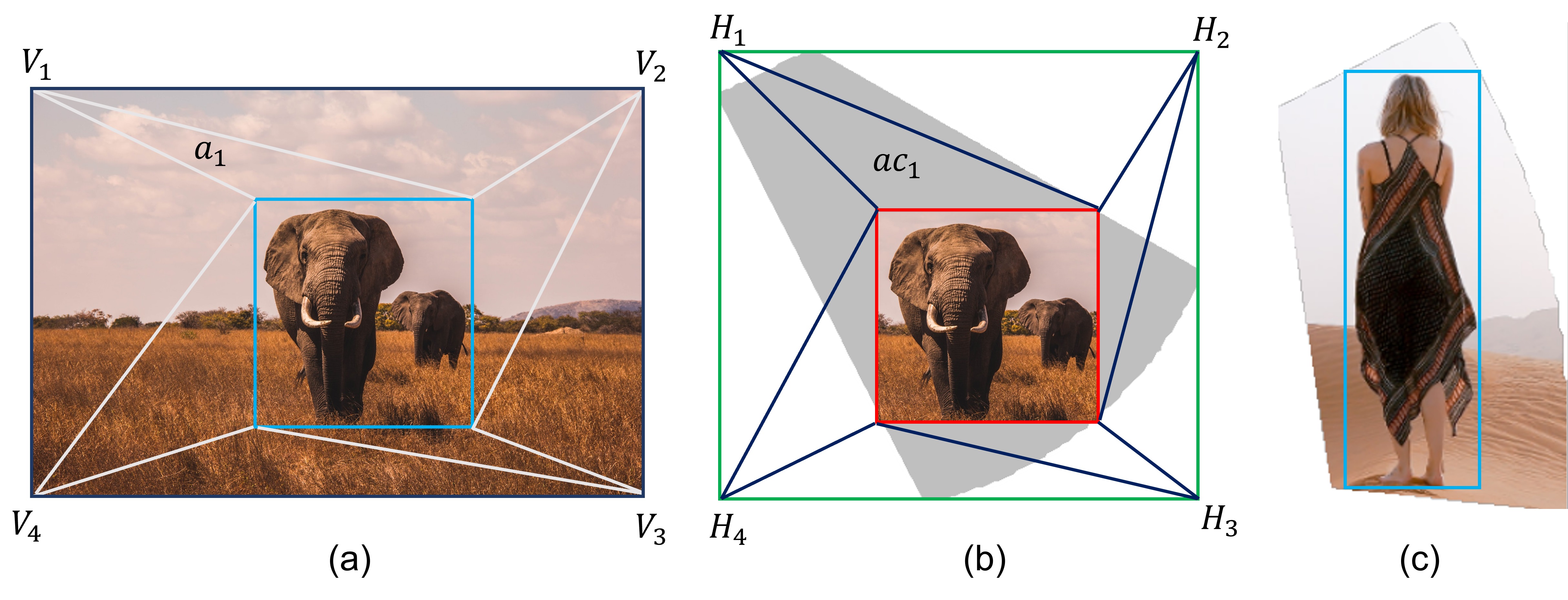}
\caption{\textcolor{black}{Two cases of filling cell. Warping triangle $a_k$ in (a) to $ac_k$ in (b). (c) is the sample case of cropping in which the bounding box is in the cell and there does not exist any empty space in the cell.}}
\label{fig:warping}
\end{figure}

\textcolor{black}{Filling the cells with the assigned image while preserving the main subjects of the image in the estimated box is the goal of this session. For example, in Fig.\ref{fig:optimize_process}, after warping, the cat in (a) is still similar to the cat in (c), but the cat's neighboring region area in (c) is warped to fill the cell. As the optimization is already successful in finding
the best fit cell for images and maximizes the area of salient box $T_i$ on the cell, a lightweight strategy can resolve the problem of filling the cell here. We consider two cases: (1) the cell is filled by image content and (2) the reverse case. For the first case, we simply crop the image along the boundary of the cell, as shown in Fig.\ref{fig:warping}-(c). For the second case, we adopt the warping of affine transformation to fill the image content to the rest of the cell. We elaborate as follows.}

\textcolor{black}{We denote the rectangle that covers an image $\mathbf{I}$ is $\mathbf{D}$ with four vertices $\mathbf{V}_1, \mathbf{V}_2, \mathbf{V}_3, \mathbf{V}_4$; and the image $\mathbf{I}$ has a bounding box $Sb_i = [bx_1, by_1, bx_2, by_2]$. We generate the delaunay triangulations for the convex hull formed by the edge of $Sb_i$ and $\mathbf{D}$, see Fig\ref{fig:warping}-(a). We denote this triangle set as $\mathbf{A}^i = \{a_k\}, k = 1, \dots, 8$. In the corresponding cell $C$ of $\mathbf{I}$, we construct a rectangle $\mathbf{H}$ (with four vertices $\mathbf{H}_1, \mathbf{H}_2, \mathbf{H}_3, \mathbf{H}_4$) that covers $C$ based on the convex vertices (Fig.\ref{fig:warping}-(b)). Similarly, we generate the delaunay triangulations of the convex hull formed by the edge of $T$ and $\mathbf{H}$. We denote this triangle set as $\mathbf{A}^c = \{ac_k\}$. To fill the image to the cell, we aim at warping $a_k$ to $ac_k$. Theoretically, the textures of pixels $p' \in ac_k$ are formulated as:
\begin{equation}
   p'(x', y') = \zeta(p(x, y)), 
\end{equation}where $p(x, y) \in a_k$, $\zeta(.)$ is the warping function of affine transformation which warps $a_k$ to $ac_k$.}

\section{Experimental Results}
\subsection{\textcolor{black}{Experiment parameters}}
\textbf{Experimental data} In our experiments, we have collected 73 different shapes and 6 image collections. The shapes are from MPEG-7 Core Experiment CE-Shape-1 Test Set \cite{latecki2000shape}, a dataset commonly used in shape research \cite{belongie2002shape, yang2016invariant, hofer2017deep}. MPEG-7 contains 1,400 shapes belonging to 70 categories. Since the shapes in each category are similar, we select one shape from every category as our testing shapes. Shapes that are unsuitable as a collage contour e.g. containing too many broken or small pieces are removed. Because most of the shapes in MPEG-7 dataset are not aesthetic and intuitive, we additionally consider 11 commonly used shapes e.g. dogs and cars. The 73 shapes are presented in the supplementary materials. For image collection, we use the AIC dataset proposed by \citet{yu2022softcollage}, which has more than 500 image collections with more than 18,000 images. The size of every collection in this dataset ranges from 10 to 100. In AIC, each image is associated with one category and one salient mask, which are useful for conducting our experiments. We only use a small subset of the AIC dataset, which is also listed in supplementary materials.

\begin{figure}
\centering
  \includegraphics[width=0.9\textwidth]{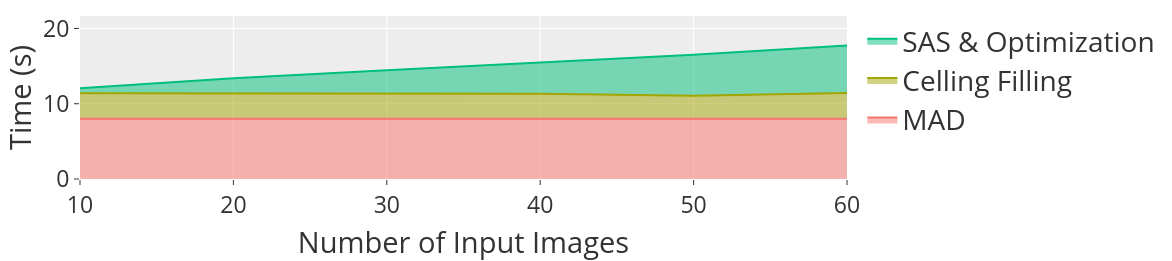}
\caption{Execution time of various steps of our method including MAD, SAS and Optimization and cell filling.}
\label{fig:time_statistics}
\end{figure}

\textbf{Implementation details} Images are first analyzed by \cite{pang2020multi} and \cite{talebi2018nima}, which usually takes one second per image on NVIDIA GTX1080Ti. The rest of our system runs on Intel Core i7-8700  with 32GB RAM. The time statistics are shown in Fig. \ref{fig:time_statistics}. The overall execution time to generate one collage ranges from 10 to 20 seconds depending on image collection sizes. Our SAS \& Optimization step takes under six seconds, which grows linearly in terms of number of input images. The other two steps i.e. MAD and cell filling in total take around 10 seconds and remain (near) constant for all image counts. To access our results and dataset, please visit our project website \href{http://graphics.csie.ncku.edu.tw/shapedimagecollage/}{\url{http://graphics.csie.ncku.edu.tw/shapedimagecollage/}}.

\subsection{Our Results and Discussion}
\textcolor{black}{
To evaluate our method, we exhibit our generated collage results in Fig.\ref{fig:comparison}. Some of these shapes have been used in prior research and commercial applications. Yet, in our study, with our SAS algorithm in generating the layout, appealing results can be generated in a balanced and visually pleasing collage. Besides, with our collaging strategy, i.e., considering both image content and the input shape in optimizing, the subjects of images can be captured and preserved well in cells. We visually show our system's ability with the competition on the results of prior works in the coming subsection.}

\begin{figure}
\centering
  \includegraphics[width=0.68\textwidth]{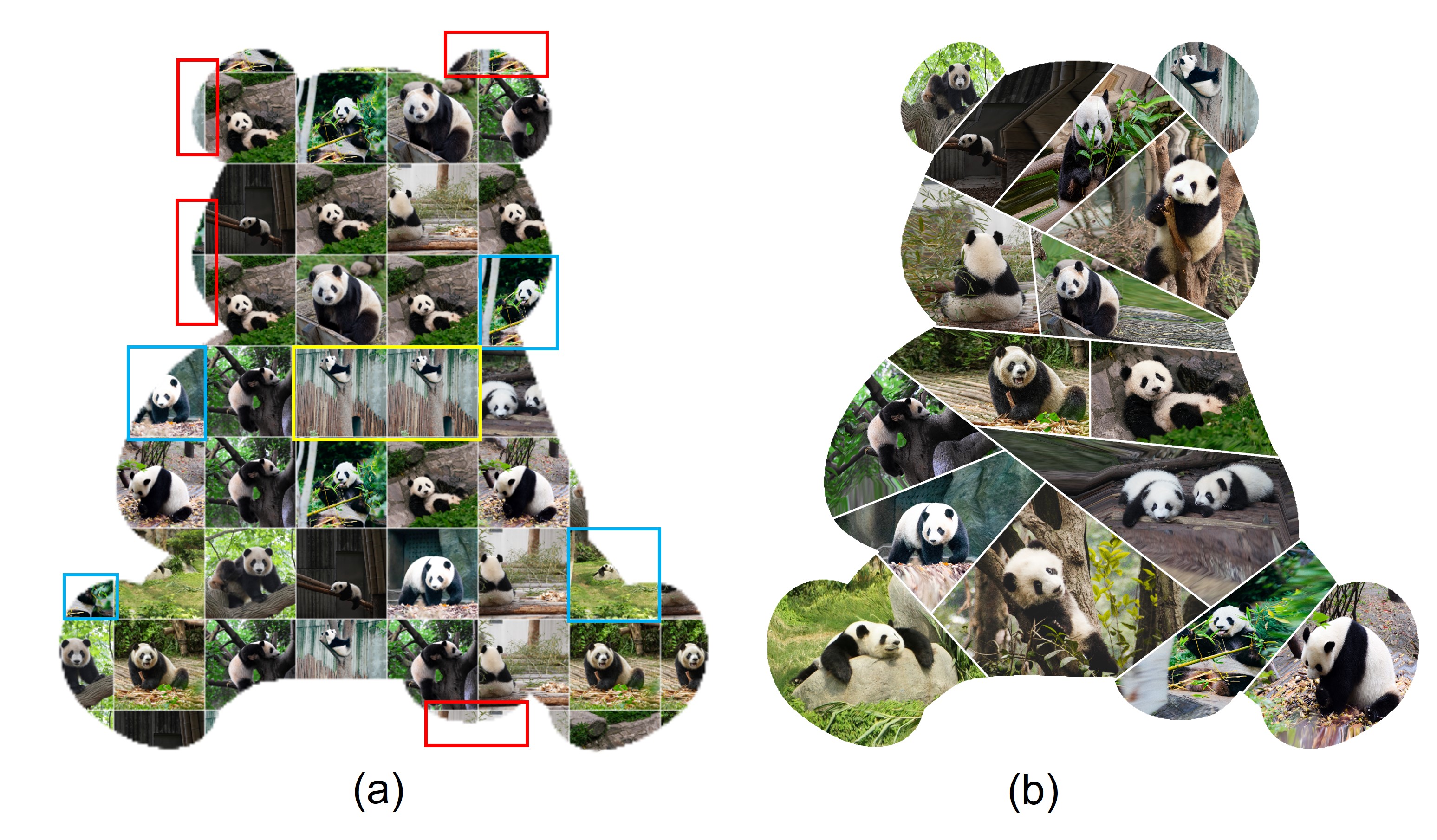}
\caption{\textcolor{black}{Visualizing the differences in collage results with layouts generated by linear slicing (a) and our SAS algorithm (b) on Panda layout.}}
\label{fig:discuss_result}
\end{figure}

\begin{figure}
\centering
  \includegraphics[width=0.67\textwidth]{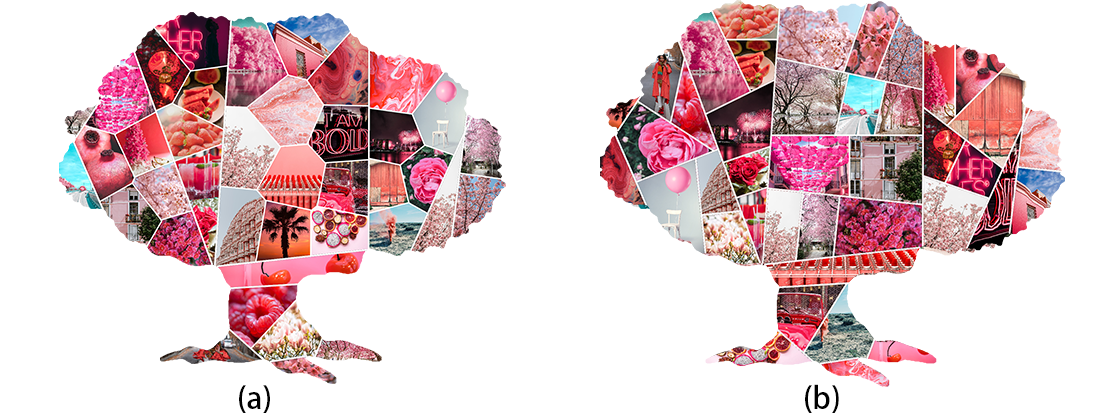}
\caption{(a) Apply centroidal Voronoid tessellation in place of SAS. (b) Using SAS.}
\label{fig:discuss_result2}
\end{figure}

\textbf{Balanced layout. }One of the interesting factors that contribute to the appealing results in this work is our proposed SAS algorithm. SAS excels in various aspects. First, generating realistic layouts with challenging shapes: let us take an example with Panda (Fig.\ref{fig:discuss_result}) as the example. The previous works linearly divide the shape into rectangles and squares, this causes the artifacts at the boundary, i.e., the boundary cells appear in form of a tiny part of other cells. That is the reason there exist several \say{useless} tiny cells surrounding the boundary as they are too small to collage meaningful content (we highlight this phenomenon in red rectangles in Fig.\ref{fig:discuss_result}-(a)). Reversely, our SAS algorithm considers the convexity and concavity of polygons when slicing the shape; thus, the generated layouts are more realistic and eliminate the \say{useless} cells. For example, the ears of the panda are well sliced and not too tiny to visualize the content in that cell. Second, the style of the cell is consistent across the layout. Since the MAD and SAS both are based on the medial axis, they have a consistent partitioning style. In contrast, if any other tessellation techniques that have no knowledge of the medial axis are used, there will be conflicting cell styles. For example, Fig. \ref{fig:discuss_result2}-(a) is the cells generated by applying centroidal Voronoi tessellation \cite{du1999centroidal} after MAD. There is a clear trace of two distinct processes i.e. the linear division style of MAD and the honeycomb style of Voronoi tessellation, whereas SAS integrates with MAD seamlessly as shown in \ref{fig:discuss_result2}-(b). Third, the number of cells can be precisely controlled and match the exact volume of the collection. This aspect is owning to being aware of the area in each well-defined region when constructing the \textit{MABSTs}. Without a clear understanding of the shape, previous methods use an indefinite number of images to fill the canvas. That is the reason there exist several images appearing many times in the resultant collage in previous work (yellow rectangles in Fig.\ref{fig:discuss_result}-(a)). In contrast, the number of cells in our generated layout is equal to the volume of the set; and thus, the collage can fully visualize the story of the given collection.

\textcolor{black}{\textbf{Semantic collage. }The major difference between our proposed scheme and prior work and commercial applications is the integration of the relation between the image content and layout structure. This enables our system to generate results in a harmonious and visually pleasing way. The balanced and visually pleasing aspects are demonstrated as the regions that attract human focus are collaged by the images with a higher interest in the collection. As shown in Fig.\ref{fig:discuss_result}, the results in the prior methods fail to connect the semantics of the collection to the layout. That is, the region highlighted in yellow is collaged by the images with the background dominating, while the images with major objects are placed at the boundary. Thus, the important objects in images are cropped in these cells (highlighted in blue rectangles). In contrast, in our results, the higher interested images are collaged in the regions which attract human focus while the landscape scenes are placed at the boundary areas. }

\textbf{Adaptive to various shapes, cell counts and sizes of image collection.} Being able to deal with arbitrary shapes consistently is challenging. For example, Voronoi tesselation is ill-defined on concave shapes. In sharp contrast, thanks to MAD, our method can decompose shapes into convex parts. The other contributing factor is our tree slicing structure. Our tree slicing structure allows us to flexibly control the number of cells and the relative size of each cell. This aspect explains why tree-based methods are standard in image collage research. However, the difference is that we generalize it to irregular canvas. Fig. \ref{fig:disuss_3} exhibits these interesting results. In particular, on the same input shape, we can generate  even-sized and uneven-sized layouts while maintaining the balance of the resultant collage. Or, also on this shape, we can produce appealing collages with different sizes of collection (e.g., 15 images and 25 images are used in this example.)

\begin{figure}
\centering
  \includegraphics[width=0.68\textwidth]{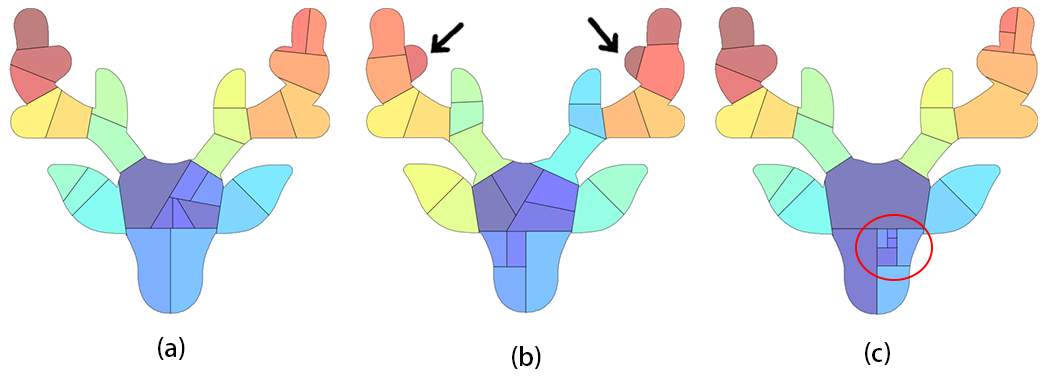}
\caption{(a) Our result using \emph{Unbalanced} $\gamma^u$ on deer shape. (b) Result when increasing the protrusion strength threshold $\tau$ to 0.9. (c) Increase $\gamma^u$ probability from 70\% to 90\%.}
\label{fig:parameter}
\end{figure}

\textbf{Effect of Parameter Settings. }
Balance of our layout is one of the aspects that affects the final collage results. To partition a given shape into a balanced layout, our scheme integrates two algorithms, MAD and SAS. Being sensitive to the different parameters in these algorithms is the issue we consider when configuring our system. More specifically, the changes in the protrusion strength threshold in MAD and the $\gamma^u$ probability in SAS have an impact on the results. Although the impact is minor in both MAD and SAS, the changes in these parameters have some visible effect on our layout generation. Fig.\ref{fig:parameter}-(a) is the result with \emph{Unbalanced} $\gamma^u$ in our SAS. Fig.\ref{fig:parameter}-(b) is the result when the protrusion strength threshold $\tau$ in MAD is increased from 0.75 to 0.9, allowing more details to be decomposed.  It can be seen that the horn details are more visible (pointed out by arrows). However, for shapes with lots of fine details, e.g. tree leaves, $\tau_p$ should be set lower to avoid an excessive amount of noise. In Fig.\ref{fig:parameter}-(c), the probability for $\gamma^u$ (in SAS) is increased from 70\% to 90\% creating higher contrast in cell sizes i.e. a large cell in the middle and tiny cells highlighted with red color. Nevertheless, users are not encouraged to set $\gamma^u$ above 90\% as it will create cells that is too small to be visible. Lastly, when choosing these parameters, users can also take into consideration the importance distribution of the image collection. For example, if the image collection has a large amount of less important images, we can use the parameters that create smaller cells, as discussed earlier, for these images.

\subsection{Ablation study}
\begin{figure}
\centering
  \includegraphics[width=0.69\textwidth]{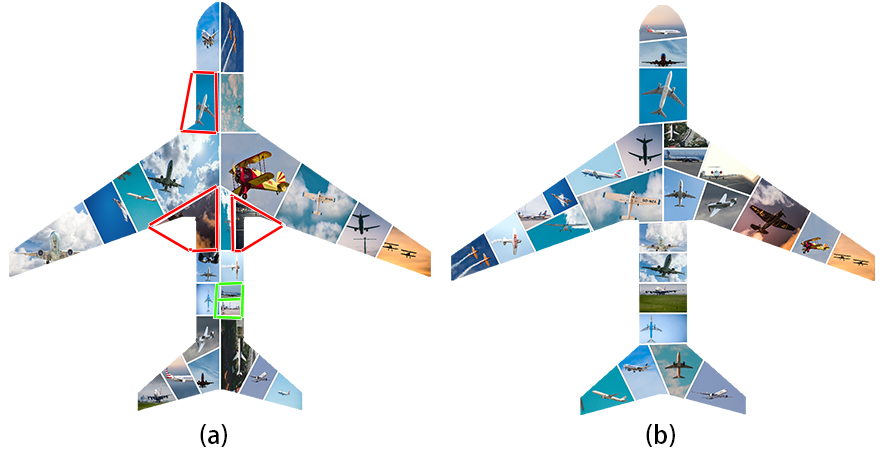}
\caption{(a) Our result without MAD. (b) Our result with MAD.}
\label{fig:ablaction_mad}
\end{figure}

\textbf{Verify the effectiveness of MAD. } We used MAD as our first step in dealing with complex shapes. We test our system without the use of MAD. The result is shown in Fig. \ref{fig:ablaction_mad}. It can be seen that many objects are heavily cropped (highlighted in red), especially in concave corners. Furthermore, without MAD we cannot precisely estimate how many images at each region. The result is that cells might end up having very different sizes. The cells highlighted in green are considerably smaller than other larger cells.

\begin{figure}
\centering
  \includegraphics[width=0.65\textwidth]{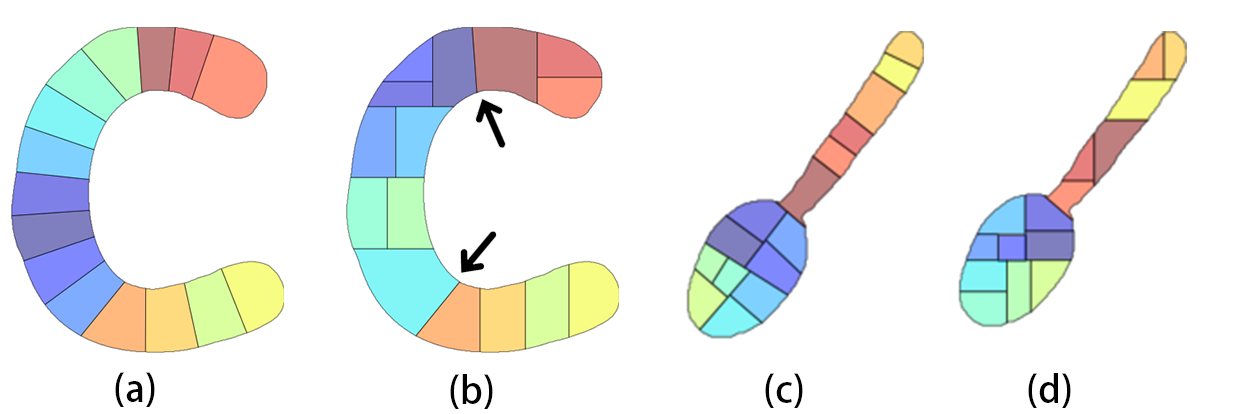}
\caption{(a) Using \emph{Axial} and \emph{Crosswise} direction on C-shape. (b) Without using \emph{Axial} and \emph{Crosswise} direction on C-shape. (c) Using \emph{Axial} and \emph{Crosswise} direction on spoon shape from MPEF7 dataset. (d) Without using \emph{Axial} and \emph{Crosswise} direction on spoon shape from MPEF7 dataset.}
\label{fig:ablaction1}
\end{figure}

\textbf{Verify Axial and Crosswise direction in SAS.}
One of the key features of SAS is the use of medial axis. We test the SAS without the use of \emph{Axial} and \emph{Crosswise} and use horizon and vertical direction instead. \emph{Balanced} strategy is used and everything else is kept the same. The difference is illustrated in Fig. \ref{fig:ablaction1}. Without using \emph{Axial} and \emph{Crosswise} direction, the algorithm has trouble finding the most intuitive way to slice the C shape and the spoon shape, resulting in cells that are less uniform in size. Moreover, it suffers from the same drawback as Voronoi tessellation i.e. different partitioning styles. For example, in Fig. \ref{fig:ablaction1}-(b), there are cuts that stand out from the rest because they are not in vertical and horizontal directions (pointed out by arrows).

\begin{figure}
\centering
\begin{subfigure}{0.22\textwidth}
  \centering
  \includegraphics[width=\textwidth]{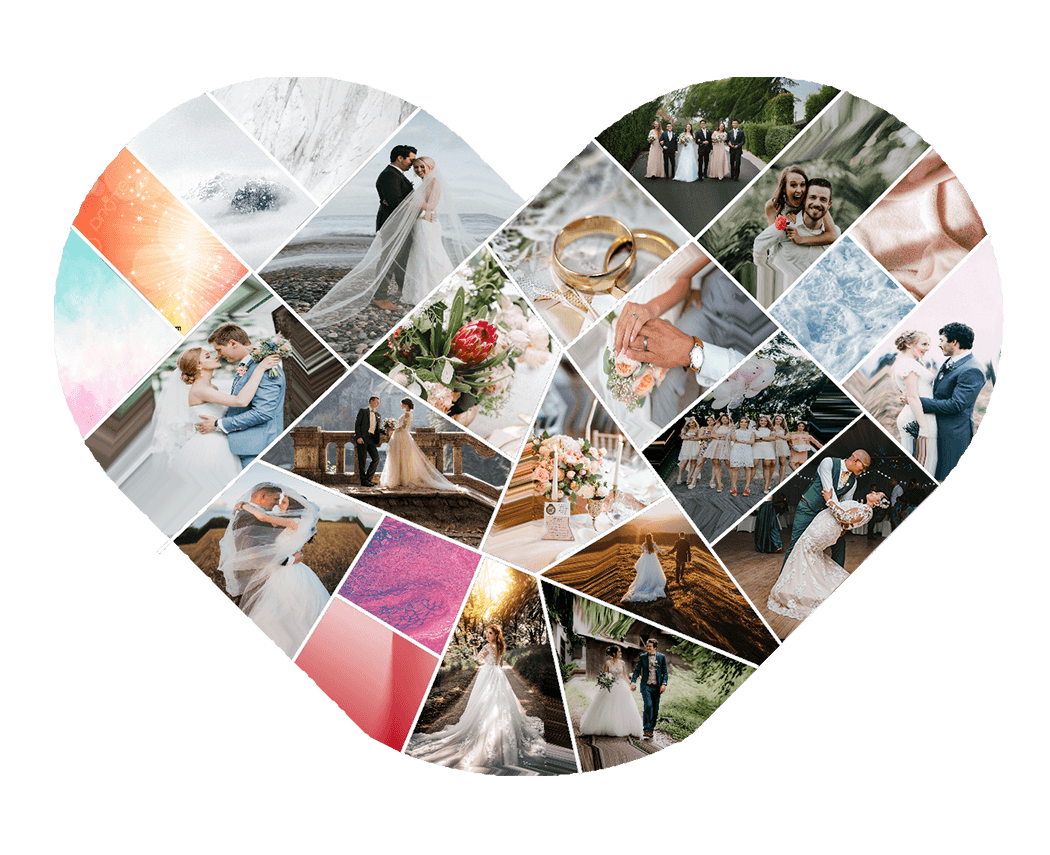}
  \caption{Result with all the components}
  \label{fig:ab_opt1}
\end{subfigure}
\begin{subfigure}{0.22\textwidth}
  \centering
  \includegraphics[width=\textwidth]{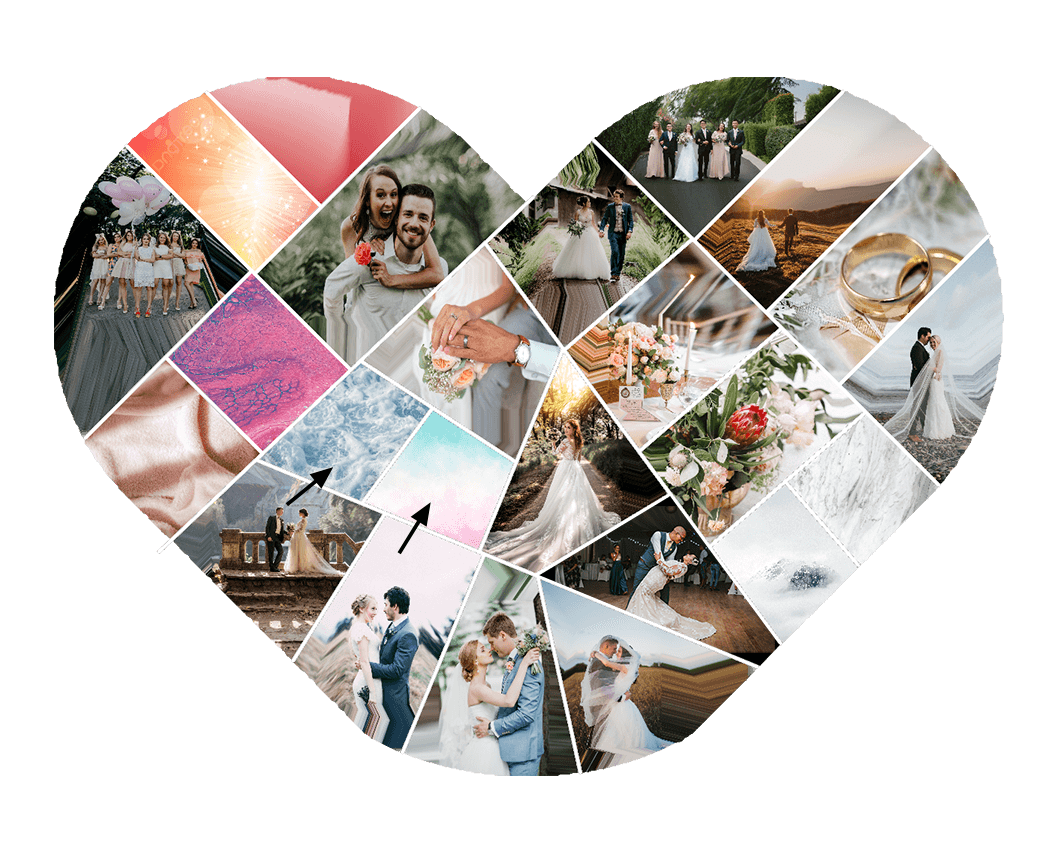}
  \caption{w/o image assignment}
  \label{fig:ab_opt2}
\end{subfigure}%
\begin{subfigure}{0.22\textwidth}
  \centering
  \includegraphics[width=\textwidth]{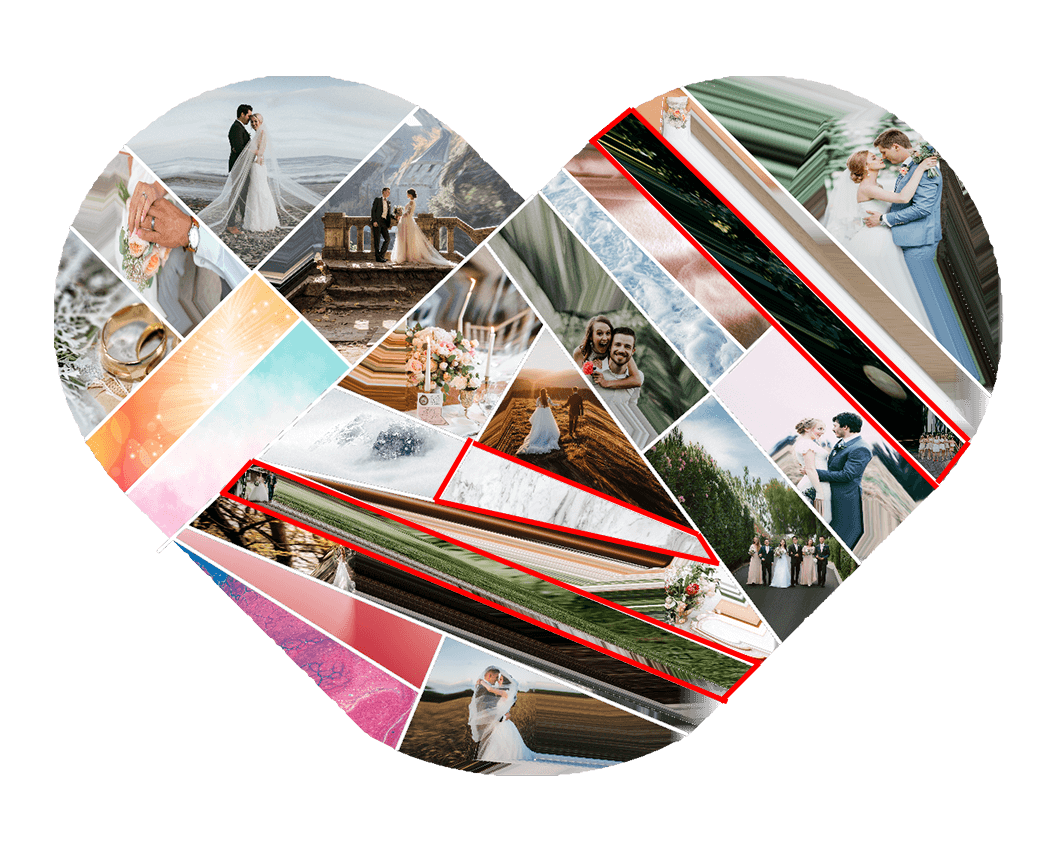}
  \caption{w/o saliency maximization}
  \label{fig:ab_opt2}
\end{subfigure}
\begin{subfigure}{0.22\textwidth}
  \centering
  \includegraphics[width=\textwidth]{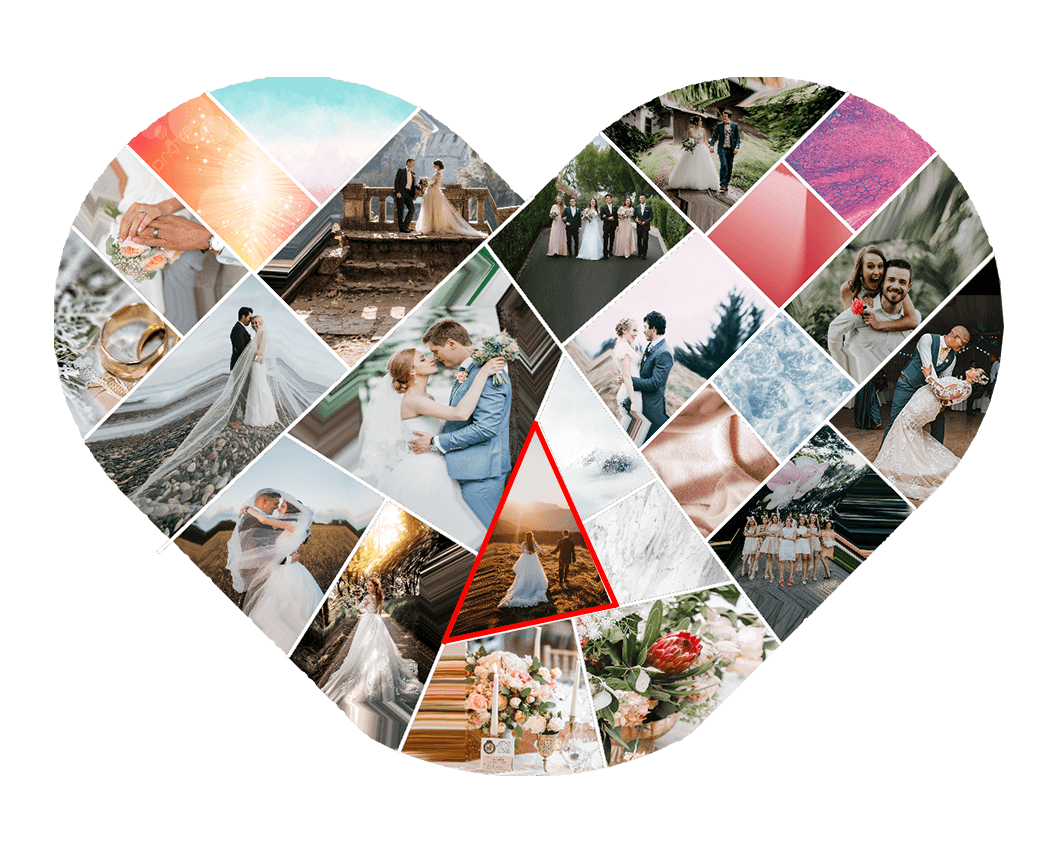}
  \caption{w/o triangle penalty}
  \label{fig:ab_opt4}
\end{subfigure}%
\caption{Visual comparison of ablated results.}
\label{fig:ablation_optimization}
\end{figure}

\begin{figure}
\centering
\begin{subfigure}{0.22\textwidth}
  \centering
  \includegraphics[width=\textwidth]{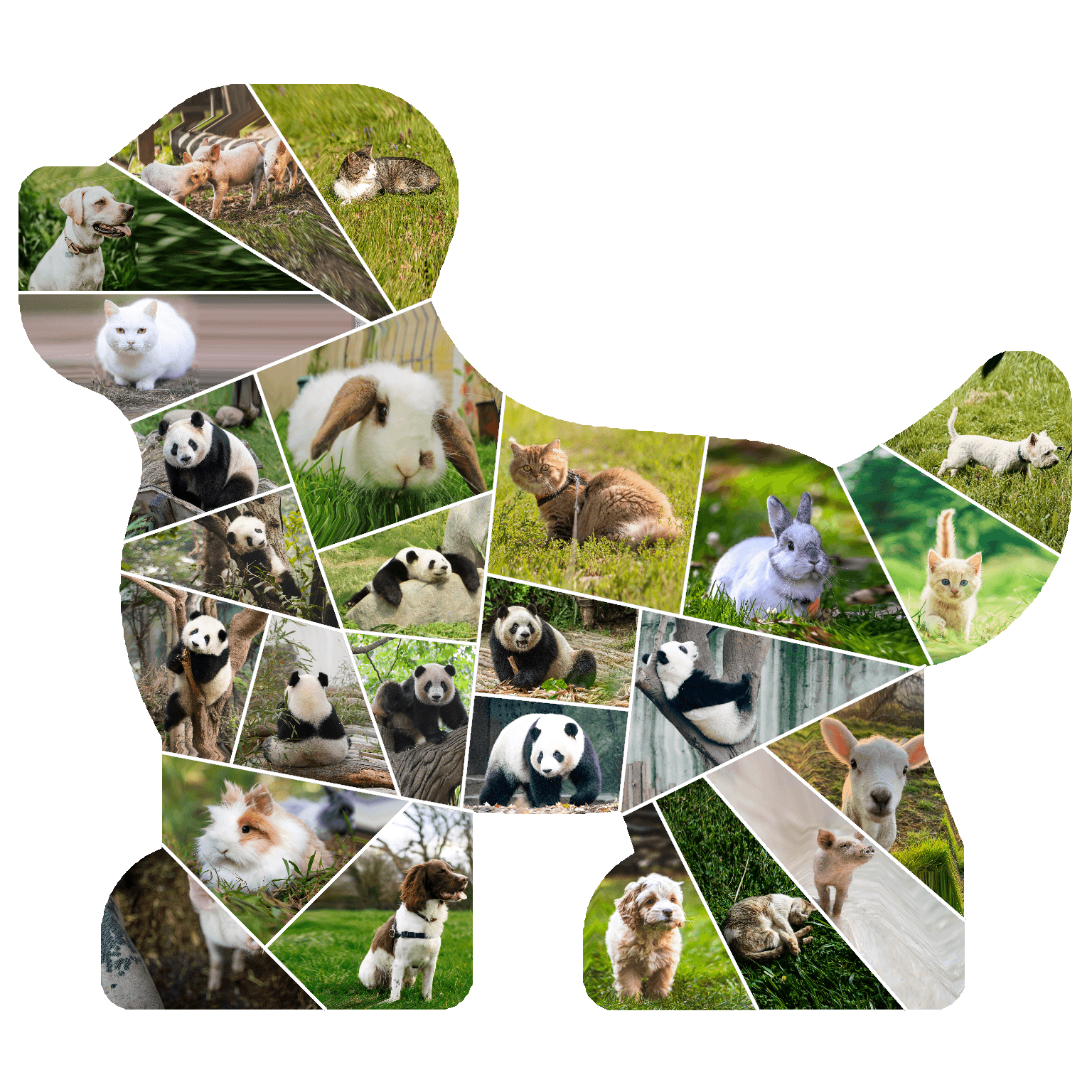}
  \caption{Even-sized layout}
  \label{fig:even}
\end{subfigure}
\begin{subfigure}{0.22\textwidth}
  \centering
  \includegraphics[width=\textwidth]{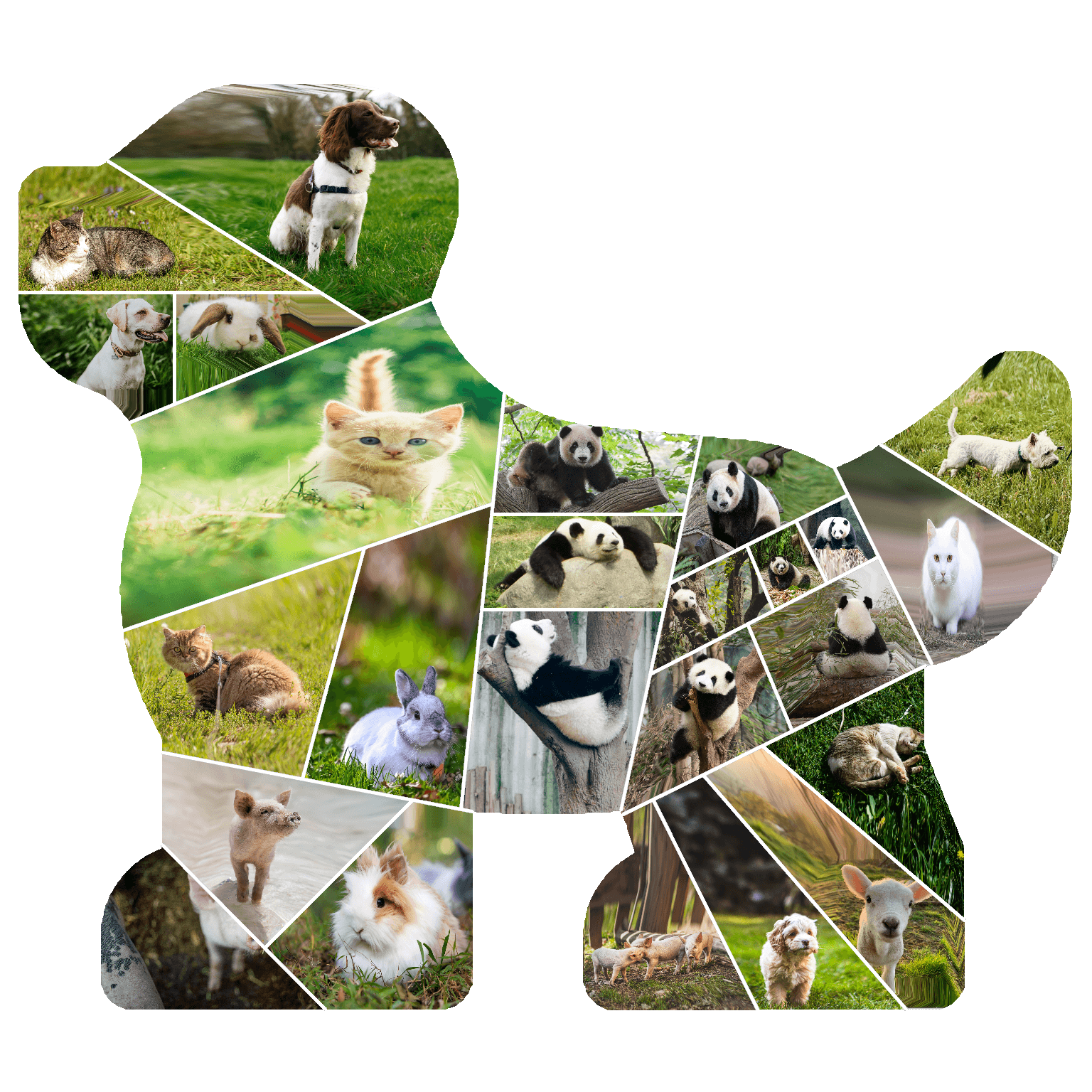}
  \caption{Uneven-sized layout}
  \label{fig:uneven}
\end{subfigure}%
\begin{subfigure}{0.22\textwidth}
  \centering
  \includegraphics[width=\textwidth]{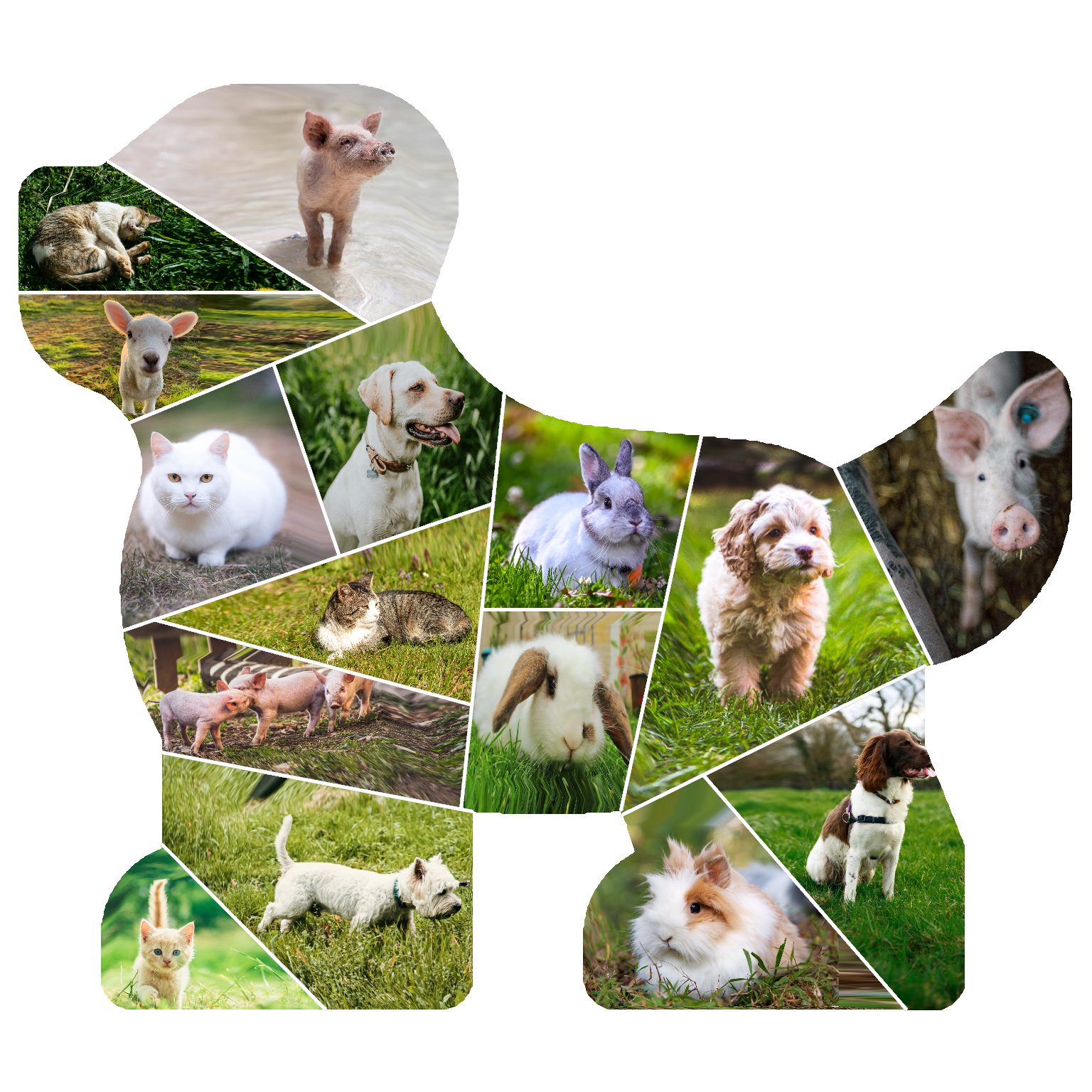}
  \caption{Collage of 15 images}
  \label{fig:image15}
\end{subfigure}
\begin{subfigure}{0.22\textwidth}
  \centering
  \includegraphics[width=\textwidth]{advanced_border_pet_extend.png}
  \caption{Collage of 25 images}
  \label{fig:image25}
\end{subfigure}%
\caption{\textcolor{black}{Demonstrates our method to be flexible in layout design and distinct sizes of image collections.}}
\label{fig:disuss_3}
\end{figure}

\textbf{Image assignment and Optimization} The image assignment step and optimization step are critical for the final quality of our results. To analyze the impact of each of them in our final results, we respectively remove each of them and compare their results with those in the full configuration. The visualization is shown in Fig. \ref{fig:ablation_optimization}. In Fig. \ref{fig:ablation_optimization}(b), we randomly assign images to leaf nodes without considering importance. Less important images might be placed in a more prominent location, in this case, background images are placed in the center (pointed out by black arrows). Our optimization has two objectives: saliency maximization and triangle penalty. The saliency maximization term simultaneously creates and matches the most suitable cells for the subjects. We create the result without optimization (i.e. randomly configure the MABST) in Fig. \ref{fig:ablation_optimization}(c). There are plenty of cells in weird shapes (highlighted in red), which are hard to place objects. Furthermore, the main subjects appear smaller in Fig.\ref{fig:ablation_optimization}(c). This means that we fail to find the tailored cells. Compared with the result in Fig.\ref{fig:ablation_optimization}(d), our method is able to suppress the triangular cell that would otherwise appear in middle (highlighted in red). Note that our method cannot always completely remove the triangles. For shapes with a curved medial axis like the heart shape, triangles are sometimes required. However, our method is able to reduce the number and the size of the triangles or at least push the triangular structure to the boundary.

\subsection{Evaluations}

\subsubsection{Qualitative Evaluation}

Here, we qualitatively evaluate the results by visually comparing our results with four baselines. The first baseline TB\cite{han2015tree} is the most related work to ours. The second method is a widely-used commercial software Shape Collage (SHP)\cite{shapecollage}. Since most of the works in image collage are on rectangular layouts, we compare with the current state-of-the-art SoftCollage(SC)\cite{yu2022softcollage}. SC only can work on rectangle layout; we further do experiment by applying shape masks to SC (SC+ Mask). Fig.\ref{fig:comparison} outlines this comparison. More comparisons are presented in the supplementary materials.

\begin{figure}
\centering
\includegraphics[width=0.99\textwidth]{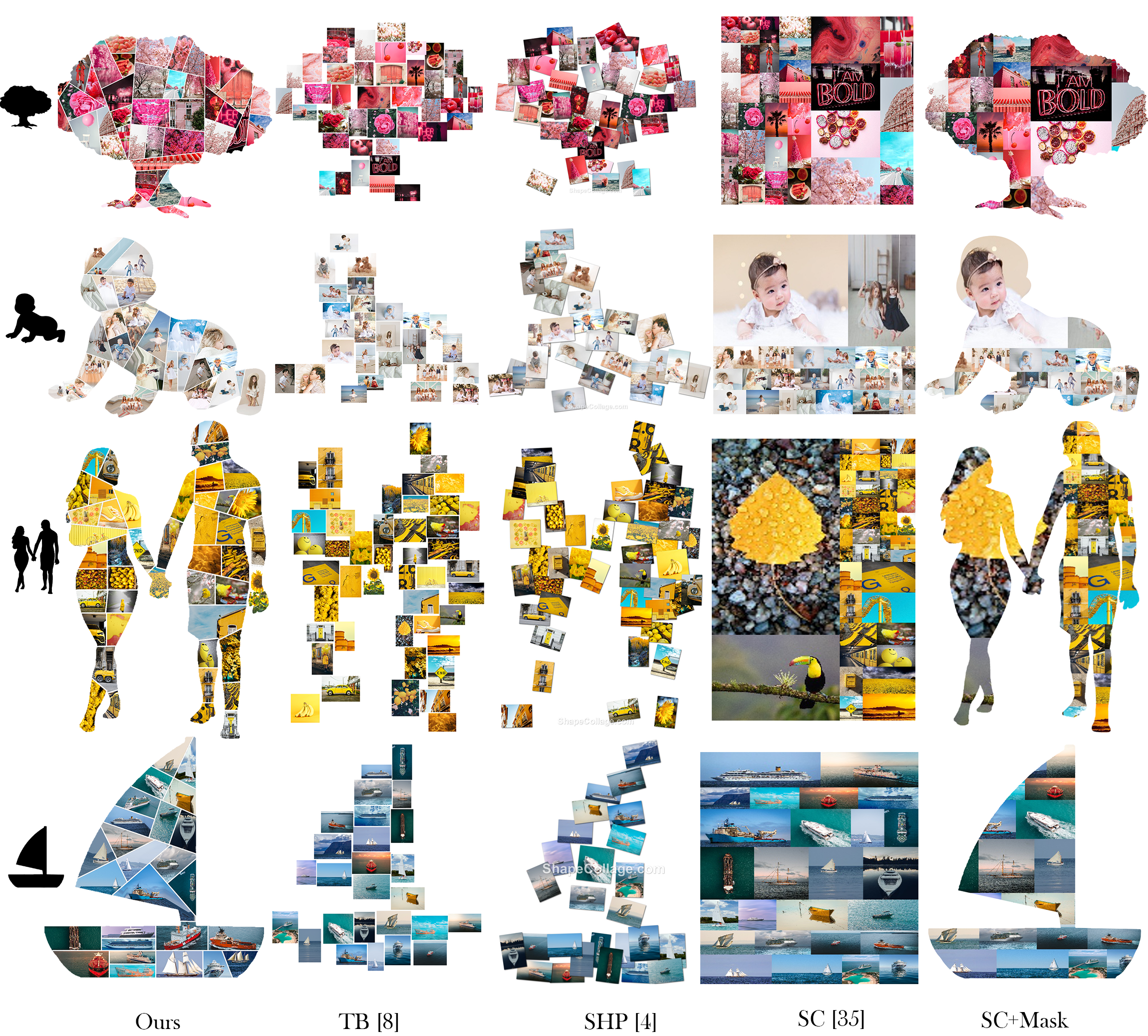}
\caption{Comparison of the results generated with different methods. The input shape are trees, babies, couples and boats, which are represented by the black silhouettes.}
\label{fig:comparison}
\end{figure}

Comparing our results with TB's \cite{han2015tree}, TB also addresses the ICAS problem. The images are first embedded in 2D canvas with hyperbolic projection, which maintains image correlations. Then they progressively adjust image locations to go within the target shapes. The adjustment process, however, is done locally and does not consider the shape as a whole. Hence, their method only works well for shapes similar to a circle. This can be seen in the examples that the tree shape in Fig. \ref{fig:comparison} works better than the others. Furthermore, their method only makes sure that the center for every image is moved inside the shape. This assumption works fine if images are tiny. But in the cases where the images are large, the majority of the image might locate outside the target region, for example in Fig. \ref{fig:comparison}, the hand of the baby or the feet of the couple. This leads to difficulties for us to recognize the shape. In contrast, it is very easy for us to recognize the shape of our results.

SHP \cite{shapecollage} is a popular image collage software that is used as the baseline model in several papers \cite{yu2022softcollage,pan2019content,han2015tree}. SHP is different from TB in that it allows for more image overlapping the image rotation. It can improve the shape accuracy in some parts, for instance, the baby's head or the woman's hair in the couple shape. However, SHP has more empty space and image overlapping that prevent it from effectively visualizing the whole story in the collection. Compared with our results, SHP suffers from the drawback similar to TB i.e. objects outside the boundaries. Furthermore, SHP cannot estimate accurately how many images in each region e.g. the leg of the man in the couple shape has no image. Meanwhile, the results shown in the first column demonstrate that our method outperforms the compared method in terms of controlling the number of images in each region.

Rectangular image collages have an advantage in preserving the complete content of the images. The state-of-the-art SC has done a good job in preserving the original aspect ratio of each image. However, compared with shaped collages, rectangular collages lack variety e.g. four examples in Fig.\ref{fig:comparison}-SC share similar visual structures and only differ in image contents. In contrast, our result is much more interesting and, as we will show in the user study, people judge that our results are more aesthetic.

In the case of SC+Mask, the results suffer the drawbacks of boundary cells, which we mentioned in the prior section. We can see several cells at the boundary in which the important objects in images are almost cut out or even not presented in the collage. Because SC+Mask does not consider the shape structure as our method does, it generates cuts that are not natural e.g. the vertical split in the middle of the tree or big images that extend beyond the baby's head.

In summary, our method outperforms the compared methods in its ability to represent the shape while preserving the content of the image collection. The image is laid out in a visually pleasing manner. All of these characteristics greatly enhance users' viewing experiences when viewing our collages.

\subsubsection{Quantitative Evaluation}

Besides the qualitative comparison, we quantitatively evaluate our proposed method.  We generate several image collages with the three baseline methods TB, SHP and SC+Mask. We do not consider SC in this experiment because it is not fair to compare some of the metrics on different layouts i.e. shaped layout and rectangular layout. The quality of results generated by these competitors is measured on five metrics that are commonly considered in the literature on image collage including the state-of-the-art SoftCollage \cite{yu2022softcollage}. Among them, nonoverlapping constraint $M_o$, correlation preservation $M_n$ and saliency loss $M_s$ are identical to \cite{yu2022softcollage}. Compactness $M_c$ is similar but generalized to irregular shapes. We further propose a new metric: saliency area $M_a$. Five metrics are described as follows:

\begin{itemize}
  
    \item \emph{Saliency area.} This metric measures the collage ability to maximize the salient objects on the canvas, which is defined as the proportion of total shape covered by salient objects. 
  \begin{equation}
    M_{a} = \dfrac{|\bigcup_{i}S_i|}{P_\mathbf{X}},
  \end{equation}.
  where $\bigcup_{i}S_i$ is the collage mask obtained by replacing each image in the collage with the corresponding saliency mask. $S_i$ is the saliency mask of image $i$. $|.|$ denotes the saliency area of the mask. $P_\mathbf{X}$ is the number of pixels of the input shape.
  
  \item \emph{Compactness.} A compact collage uses space less wastefully by minimizing white space. We formulate the compactness as:
  \begin{equation}
      M_{c} = \dfrac{P_w}{P_\mathbf{X}},
  \end{equation}
  where $P_w$ is the number of pixels of the white space.
  
  \item \emph{Non-overlapping constraint.} Image overlapping decreases the aesthetics and informativeness of the collage. Overlapping can be calculated as
  \begin{equation}
   M_{o} = \dfrac{P_o}{P_\mathbf{X}},   
  \end{equation}
  where $P_o$ is the sum of the intersecting pixels of any two images.
  
  \item \emph{Correlation preservation.} Placing correlated images together can facilitate the informativeness of the collage. The metric is expressed as: 
  \begin{equation}
    M_{n} =\dfrac{1}{N}\sum_{i} \|(L_i - L_{ci})\|,
  \end{equation}
 where $L_i$ is the location of image $i$ in the collage, and $L_{ci}$ is the location of the centroid location of the category $ci$ of image $i$, which are provided in AIC dataset. For this metric, the lower is better. All location coordinates are normalized by the width and height of the input shape.
 
  \item \emph{Saliency loss.} This metric measures the ability to preserve salient regions in the image and is defined as 
  \begin{equation}
    M_{s} = 1 - \dfrac{|\bigcup_{i}S_i|}{\sum_{i}|S_i|}.
  \end{equation}
\end{itemize}

\begin{table}[h!]
\centering
\caption{Quantitative Evaluation Metrics}
\begin{tabular}{c c c c c c} 
 \hline
 Method & $M_a$ & $M_c$ & $M_o$ & $M_n$& $M_s$\\ [0.5ex] 
 \hline
    TB\cite{han2015tree} & 0.08 & 0.23 & 0.01 & \textbf{0.12} & \textbf{0}  \\ 
    SHP\cite{shapecollage}& 0.12 & 0.29 & 0.09 & 0.15& 0.06  \\
    SC+Mask & 0.19& \textbf{0} & \textbf{0} & 0.13 & 0.52\\
    Ours& \textbf{0.32} & \textbf{0} & \textbf{0} & 0.17 & \textbf{0} \\
 \hline
\end{tabular}

\label{table:quantitative_evaluation}
\end{table}

Table \ref{table:quantitative_evaluation} shows the statistic on the above evaluation metrics. For the first metric $M_a$ the higher is better. For all the other metrics, the lower is better. The first thing to notice is that our method achieves the higher in the first and the lowest value in three of the other metrics i.e. $M_c$, $M_o$, $M_s$, while performing similarly to the competitors in $M_n$. Larger saliency area $M_a$ means that our method uses the shaped space more efficiently. Better compactness (lower $M_c$) reflects our main goal to authentically represent the input shape. Although SC+Mask also achieves zero in this metric, it lags far behind our method in $M_s$ because it is not originally designed for shaped collage. As for non-overlapping constraint $M_o$, SHP performs the worst because SHP allows for overlapping. For correlation preservation $M_n$, TB and SC+Mask beat our method and SHP due to their inclusion of image feature extraction components. However, the difference is not huge. In summary, the three baseline methods all have obvious drawbacks. For TB and SHP, the weaknesses are compactness $M_c$. For SC+Mask, the weak point is saliency loss $M_s$. This reveals that our method is the best among these four methods.

\subsection{User study}
We conduct two user studies to evaluate the effectiveness of our results. One is to measure users' preference for different methods, and the other is to measure how effective is our method in presenting the information. 16 image collections with the number of images ranging from 15 to 40 are used along with 16 different shapes. For each image collection and shape, we generate results with our methods and four baseline methods i.e. TB, SHP, SC and SC+Mask. We recruited a total of 39 users to conduct our user study. They are of different ages (age range of 21-31) and backgrounds (13 of them have graphics-related backgrounds). In the first user study, the users are asked to choose between two results generated with two of the five methods. The result of the side-by-side evaluation is shown in Table \ref{table:user_study}. In the side-by-side evaluation, our method beats all the comparative methods by 84\%, 83\%, 60\%, and 43\% respectively. The statistics results reveal that our results receive major votes from the users. It demonstrates that our method can catch the general public users' interest. The evaluation results are presented in Fig.5 of the supplementary file. When analyzing the evaluation results, we found that the examples R5 and R11 receive relatively fewer votes than other samples. It is because the layout generated by these shapes consists of some narrow regions. Thus, they could not be favored by the users. In the second user study, users are given a collage result and four pictures of salient objects that appear in that collage. We measure the total time for the user to locate all four objects in the collage. Our result has the second lowest retrieval time among the five methods as shown in Table \ref{table:user_study2}. The SC+Mask achieves the lowest time because it has far fewer objects to check compared to the others. We can conclude that our method can effectively present the data, which allows users to easily consume the information.

\begin{table}[H]
\centering
\caption{Side-by-side User Evaluation.}
\begin{tabular}{c c c c c} 
 \hline
 & Wins & Equally Good & Losses &$\Delta$\\ [0.5ex] 
 \hline
    Ours v.s. TB\cite{han2015tree} & 91\% & 2\% & 7\% & \textbf{84\%}\\ 

    Ours v.s. SHP\cite{shapecollage} &90\% & 3\% & 7\% & \textbf{83\%}\\

    Ours v.s. SC\cite{yu2022softcollage}& 77\% & 6\% & 17\% & \textbf{60\%}\\

    Ours v.s. SC+Mask& 69\% & 5\% & 26\% & \textbf{43\%}\\
 \hline
\end{tabular}

 $\Delta$ denotes the difference of the win rate and the loss rate. Higher is better.
\label{table:user_study}
\end{table}

\begin{table}[h!]
\centering
\caption{Information Conveying Test.}

\begin{tabular}{c c c c c c} 
  & Ours & TB\cite{han2015tree} & SHP\cite{shapecollage} &SC\cite{yu2022softcollage} & SC+Mask\\ [0.5ex] 
  \hline
  Time (s) & 18.15 & 20.03 & 22.25 & 18.47 & 13.40\\ 

\end{tabular}
\label{table:user_study2}
\end{table}

\subsection{Limitation}
\begin{figure}[H]
\centering
  \includegraphics[width=0.45\textwidth]{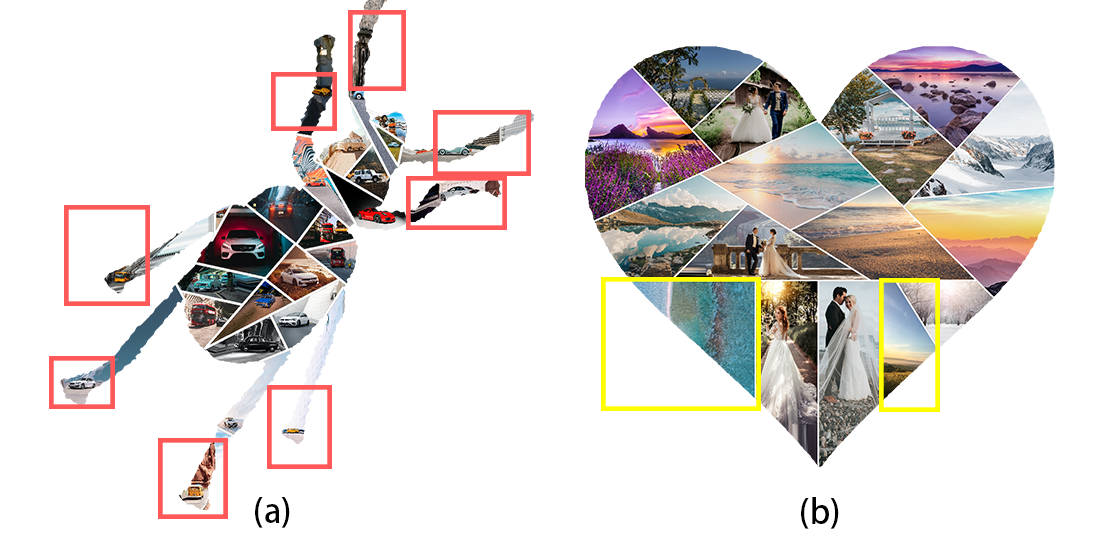}
\caption{\textcolor{black}{Two examples of limitations of our system. In (a), the beetle shape consists of multiple narrow regions. This leads to small images (highlighted in red). In (b), the number of landscape photos dominates in the given collection. We may not analyze the subjects of such photos precisely. And thus, the optimization step may fail to estimate the correct cell and the box in the cell to put such landscape photos. That is the reason, the scene of several images is cropped (highlighted in yellow). This eventually damages the semantic and visually pleasing factors of the final collage.}}
\label{fig:limitation}
\end{figure}

\textcolor{black}{We have presented a system that collages various image collections in diverse shapes. However, for shapes that have very long and narrow regions (illustrated in Fig. \ref{fig:limitation}-(a), our method can work but the visual quality of the result is less than ideal. In particular, the image content in the legs of the beetle is not identifiable. This stems from our problem formulation. A different formulation might be better to deal with this case e.g. collage on the complement of the beetle shape.} Another limitation occurs when the image collection encompasses landscape photos, our method may not perform well (as shown in Fig.\ref{fig:limitation}-(b)). Currently, our approach adopts an off-the-shelf salient object detection method, which is introduced in \cite{pang2020multi}, to detect the subjects in images. In the landscape photos, the difference in salient values in patches is small. Therefore, our optimization scheme may fail to estimate the tailored cell and target box to collage such photos.

\section{Conclusion}
In this paper, we introduce a unified ICAS algorithm centered around medial axis. The algorithm includes a novel Shape-aware Slicing algorithm and an optimal collage search strategy. We demonstrate that the proposed slicing method is especially useful for balancing the layout of image collage on irregular shapes. This gives our system the capability of collaging image collection with flexible and diverse shapes. Moreover, the proposed layout optimization serves better collages by analyzing the correlation between the content in the collection and the layout structure. Our results and evaluation show that the proposed collage scheme substantially outperforms prior works and overcomes the drawbacks in existing commercial applications. In the future, we plan to investigate such techniques to assess the semantics in the landscape photos to improve the accuracy of the optimization and thus enhance the visual quality of generated results. Furthermore, we may consider different visualization techniques for shapes with long narrow regions.

\bibliographystyle{abbrv}
\bibliography{references}

\begin{thebibliography}{36}
\providecommand{\natexlab}[1]{#1}
\providecommand{\url}[1]{\texttt{#1}}
\expandafter\ifx\csname urlstyle\endcsname\relax
  \providecommand{\doi}[1]{doi: #1}\else
  \providecommand{\doi}{doi: \begingroup \urlstyle{rm}\Url}\fi

\bibitem[Adobe(2021)]{Adobe}
Adobe.
\newblock Photo collage.
\newblock Available: \url{https://www.adobe.com/express/create/photo-collage},
  2021.

\bibitem[Atkins(2008)]{atkins2008blocked}
C.~B. Atkins.
\newblock Blocked recursive image composition.
\newblock In \emph{Proceedings of the 16th ACM international conference on
  Multimedia}, pages 821--824, 2008.

\bibitem[Belongie et~al.(2002)Belongie, Malik, and Puzicha]{belongie2002shape}
S.~Belongie, J.~Malik, and J.~Puzicha.
\newblock Shape matching and object recognition using shape contexts.
\newblock \emph{IEEE transactions on pattern analysis and machine
  intelligence}, 24\penalty0 (4):\penalty0 509--522, 2002.

\bibitem[Cheung(2013)]{shapecollage}
V.~Cheung.
\newblock Shape collage.
\newblock Available: \url{http://www.shapecollage.com/}, 2013.

\bibitem[Choi et~al.(1997)Choi, Choi, and Moon]{choi1997mathematical}
H.~I. Choi, S.~W. Choi, and H.~P. Moon.
\newblock Mathematical theory of medial axis transform.
\newblock \emph{pacific journal of mathematics}, 181\penalty0 (1):\penalty0
  57--88, 1997.

\bibitem[De~Winter and Wagemans(2006)]{de2006segmentation}
J.~De~Winter and J.~Wagemans.
\newblock Segmentation of object outlines into parts: A large-scale integrative
  study.
\newblock \emph{Cognition}, 99\penalty0 (3):\penalty0 275--325, 2006.

\bibitem[Du et~al.(1999)Du, Faber, and Gunzburger]{du1999centroidal}
Q.~Du, V.~Faber, and M.~Gunzburger.
\newblock Centroidal voronoi tessellations: Applications and algorithms.
\newblock \emph{SIAM review}, 41\penalty0 (4):\penalty0 637--676, 1999.

\bibitem[Han et~al.(2015)Han, Zhang, Lin, Xu, Sheng, and Mei]{han2015tree}
X.~Han, C.~Zhang, W.~Lin, M.~Xu, B.~Sheng, and T.~Mei.
\newblock Tree-based visualization and optimization for image collection.
\newblock \emph{IEEE Transactions on Cybernetics}, 46\penalty0 (6):\penalty0
  1286--1300, 2015.

\bibitem[Hofer et~al.(2017)Hofer, Kwitt, Niethammer, and Uhl]{hofer2017deep}
C.~Hofer, R.~Kwitt, M.~Niethammer, and A.~Uhl.
\newblock Deep learning with topological signatures.
\newblock \emph{Advances in neural information processing systems}, 30, 2017.

\bibitem[Hoffman and Richards(1984)]{hoffman1984parts}
D.~D. Hoffman and W.~A. Richards.
\newblock Parts of recognition.
\newblock \emph{Cognition}, 18\penalty0 (1-3):\penalty0 65--96, 1984.

\bibitem[Hoffman and Singh(1997)]{hoffman1997salience}
D.~D. Hoffman and M.~Singh.
\newblock Salience of visual parts.
\newblock \emph{Cognition}, 63\penalty0 (1):\penalty0 29--78, 1997.

\bibitem[Latecki and Lak{\"a}mper(1999)]{latecki1999convexity}
L.~J. Latecki and R.~Lak{\"a}mper.
\newblock Convexity rule for shape decomposition based on discrete contour
  evolution.
\newblock \emph{Computer Vision and Image Understanding}, 73\penalty0
  (3):\penalty0 441--454, 1999.

\bibitem[Latecki et~al.(2000)Latecki, Lakamper, and Eckhardt]{latecki2000shape}
L.~J. Latecki, R.~Lakamper, and T.~Eckhardt.
\newblock Shape descriptors for non-rigid shapes with a single closed contour.
\newblock In \emph{Proceedings IEEE Conference on Computer Vision and Pattern
  Recognition. CVPR 2000 (Cat. No. PR00662)}, volume~1, pages 424--429. IEEE,
  2000.

\bibitem[Lekschas et~al.(2020)Lekschas, Zhou, Chen, Gehlenborg, Bach, and
  Pfister]{lekschas2020generic}
F.~Lekschas, X.~Zhou, W.~Chen, N.~Gehlenborg, B.~Bach, and H.~Pfister.
\newblock A generic framework and library for exploration of small multiples
  through interactive piling.
\newblock \emph{IEEE Transactions on Visualization and Computer Graphics},
  27\penalty0 (2):\penalty0 358--368, 2020.

\bibitem[Lien and Amato(2006)]{lien2006approximate}
J.-M. Lien and N.~M. Amato.
\newblock Approximate convex decomposition of polygons.
\newblock \emph{Computational Geometry}, 35\penalty0 (1-2):\penalty0 100--123,
  2006.

\bibitem[Liu et~al.(2017{\natexlab{a}})Liu, Zhang, Jing, Guo, Chen, and
  Wang]{liu2017correlation}
L.~Liu, H.~Zhang, G.~Jing, Y.~Guo, Z.~Chen, and W.~Wang.
\newblock Correlation-preserving photo collage.
\newblock \emph{IEEE transactions on visualization and computer graphics},
  24\penalty0 (6):\penalty0 1956--1968, 2017{\natexlab{a}}.

\bibitem[Liu et~al.(2017{\natexlab{b}})Liu, Wang, Li, and
  Noh]{liu2017trcollage}
S.~Liu, X.~Wang, P.~Li, and J.~Noh.
\newblock Trcollage: efficient image collage using tree-based layer reordering.
\newblock In \emph{2017 International Conference on Virtual Reality and
  Visualization (ICVRV)}, pages 454--455. IEEE, 2017{\natexlab{b}}.

\bibitem[Luo et~al.(2014)Luo, Shen, Liu, and Zhang]{luo2014computational}
L.~Luo, C.~Shen, X.~Liu, and C.~Zhang.
\newblock A computational model of the short-cut rule for 2d shape
  decomposition.
\newblock \emph{IEEE Transactions on Image Processing}, 24\penalty0
  (1):\penalty0 273--283, 2014.

\bibitem[Nguyen and Worring(2008)]{nguyen2008interactive}
G.~P. Nguyen and M.~Worring.
\newblock Interactive access to large image collections using similarity-based
  visualization.
\newblock \emph{Journal of Visual Languages \& Computing}, 19\penalty0
  (2):\penalty0 203--224, 2008.

\bibitem[Ogniewicz and Ilg(1992)]{ogniewicz1992voronoi}
R.~L. Ogniewicz and M.~Ilg.
\newblock Voronoi skeletons: theory and applications.
\newblock In \emph{CVPR}, volume~92, pages 63--69, 1992.

\bibitem[Pan et~al.(2019)Pan, Tang, Dong, Ma, Meng, Huang, Lee, and
  Xu]{pan2019content}
X.~Pan, F.~Tang, W.~Dong, C.~Ma, Y.~Meng, F.~Huang, T.-Y. Lee, and C.~Xu.
\newblock Content-based visual summarization for image collections.
\newblock \emph{IEEE Transactions on Visualization and Computer Graphics},
  27\penalty0 (4):\penalty0 2298--2312, 2019.

\bibitem[Pang et~al.(2020)Pang, Zhao, Zhang, and Lu]{pang2020multi}
Y.~Pang, X.~Zhao, L.~Zhang, and H.~Lu.
\newblock Multi-scale interactive network for salient object detection.
\newblock In \emph{Proceedings of the IEEE/CVF conference on computer vision
  and pattern recognition}, pages 9413--9422, 2020.

\bibitem[Papanelopoulos et~al.(2019)Papanelopoulos, Avrithis, and
  Kollias]{papanelopoulos2019revisiting}
N.~Papanelopoulos, Y.~Avrithis, and S.~Kollias.
\newblock Revisiting the medial axis for planar shape decomposition.
\newblock \emph{Computer Vision and Image Understanding}, 179:\penalty0 66--78,
  2019.

\bibitem[Reasyze()]{shapex}
Reasyze.
\newblock Shapex.
\newblock Available: \url{https://www.reasyze.com/shapex/}.

\bibitem[Rother et~al.(2006)Rother, Bordeaux, Hamadi, and
  Blake]{rother2006autocollage}
C.~Rother, L.~Bordeaux, Y.~Hamadi, and A.~Blake.
\newblock Autocollage.
\newblock \emph{ACM transactions on graphics (TOG)}, 25\penalty0 (3):\penalty0
  847--852, 2006.

\bibitem[SilkenMermaid()]{figrcollage}
SilkenMermaid.
\newblock Figrcollage.
\newblock Available: \url{https://www.figrcollage.com/}.

\bibitem[Singh and Hoffman(2001)]{singh2001part}
M.~Singh and D.~D. Hoffman.
\newblock Part-based representations of visual shape and implications for
  visual cognition.
\newblock In \emph{Advances in psychology}, volume 130, pages 401--459.
  Elsevier, 2001.

\bibitem[Singh et~al.(1999)Singh, Seyranian, and Hoffman]{singh1999parsing}
M.~Singh, G.~D. Seyranian, and D.~D. Hoffman.
\newblock Parsing silhouettes: The short-cut rule.
\newblock \emph{Perception \& Psychophysics}, 61\penalty0 (4):\penalty0
  636--660, 1999.

\bibitem[Song et~al.(2021)Song, Tang, Dong, Huang, Lee, and
  Xu]{song2021balance}
Y.~Song, F.~Tang, W.~Dong, F.~Huang, T.-Y. Lee, and C.~Xu.
\newblock Balance-aware grid collage for small image collections.
\newblock \emph{IEEE Transactions on Visualization and Computer Graphics},
  2021.

\bibitem[Talebi and Milanfar(2018)]{talebi2018nima}
H.~Talebi and P.~Milanfar.
\newblock Nima: Neural image assessment.
\newblock \emph{IEEE transactions on image processing}, 27\penalty0
  (8):\penalty0 3998--4011, 2018.

\bibitem[Tan et~al.(2011)Tan, Song, Liu, and Xie]{tan2011imagehive}
L.~Tan, Y.~Song, S.~Liu, and L.~Xie.
\newblock Imagehive: Interactive content-aware image summarization.
\newblock \emph{IEEE Computer Graphics and Applications}, 32\penalty0
  (1):\penalty0 46--55, 2011.

\bibitem[Wong and Liu(1986)]{1586075}
D.~Wong and C.~Liu.
\newblock A new algorithm for floorplan design.
\newblock In \emph{23rd ACM/IEEE Design Automation Conference}, pages 101--107,
  1986.
\newblock \doi{10.1109/DAC.1986.1586075}.

\bibitem[Wu and Aizawa(2013)]{wu2013picwall}
Z.~Wu and K.~Aizawa.
\newblock Picwall: Photo collage on-the-fly.
\newblock In \emph{2013 Asia-Pacific Signal and Information Processing
  Association Annual Summit and Conference}, pages 1--10. IEEE, 2013.

\bibitem[Yang et~al.(2016)Yang, Wang, Yuan, Li, and Liu]{yang2016invariant}
J.~Yang, H.~Wang, J.~Yuan, Y.~Li, and J.~Liu.
\newblock Invariant multi-scale descriptor for shape representation, matching
  and retrieval.
\newblock \emph{Computer Vision and Image Understanding}, 145:\penalty0 43--58,
  2016.

\bibitem[Yu et~al.(2022)Yu, Chen, Zhang, and Li]{yu2022softcollage}
J.~Yu, L.~Chen, M.~Zhang, and M.~Li.
\newblock Softcollage: A differentiable probabilistic tree generator for image
  collage.
\newblock In \emph{Proceedings of the IEEE/CVF Conference on Computer Vision
  and Pattern Recognition}, pages 3729--3738, 2022.

\bibitem[Zeng et~al.(2008)Zeng, Lakaemper, Yang, and Li]{zeng20082d}
J.~Zeng, R.~Lakaemper, X.~Yang, and X.~Li.
\newblock 2d shape decomposition based on combined skeleton-boundary features.
\newblock In \emph{International Symposium on Visual Computing}, pages
  682--691. Springer, 2008.

\end{thebibliography}
\end{document}